%% file: prte_neurips_2024.tex
\documentclass{article}

\usepackage[preprint]{neurips_2024}
\setcitestyle{numbers,square}

\usepackage[utf8]{inputenc} 
\usepackage[T1]{fontenc}    
\usepackage{hyperref}       
\usepackage{url}            
\usepackage{booktabs}       
\usepackage{amsfonts}       
\usepackage{nicefrac}       
\usepackage{microtype}      
\usepackage{xcolor}         

\usepackage{wrapfig}

\input{tex/dependencies.tex}

\title{Probabilistic Regular Tree Priors for Scientific Symbolic Reasoning}

\author{
  Tim Schneider$^{1}$\quad Amin Totounferoush$^{1}$\quad Wolfgang Nowak$^{2}$\quad Steffen Staab$^{1,3}$\\
$^1$ Institute for Artificial Intelligence, University of Stuttgart\\
$^2$ Stochastic Modelling of Hydrosystems, University of Stuttgart\\
$^3$ University of Southhampton \\
\texttt{\{tim.schneider, amin.totounferoush,}\\ \texttt{steffen.staab\}@ki.uni-stuttgart.de}\\
\texttt{wolfgang.nowak@iws.uni-stuttgart.de}
}

\input{sections/glossary.tex}

\begin{document}

\maketitle

\input{sections/abstract.tex}
\glsresetall
\input{sections/introduction.tex}

\glsresetall
\input{sections/background}

\input{sections/related}

\input{sections/method.tex}

\input{sections/boolean.tex}

\input{sections/application.tex}
\input{sections/conclusion.tex}

\input{formal/acknowledgement.tex}

\clearpage

\bibliography{references}
\bibliographystyle{plainnat}

\newpage
\appendix
\onecolumn
\section{Implementation Details}
\label{sec:appendix:implementation_details}

\subsection{Determinant of differentiable jumps between parameter spaces}
\label{sec:appendix:logdetjacobian}
\input{sections/appendix/implementation/logdetjacobian.tex}

\subsection{Hyperparameters}
\input{sections/appendix/implementation/hyperparameters.tex}

\subsection{Context Re-Sampling}
\label{sec:appendix:context_resampling}
\input{sections/appendix/implementation/context.tex}

\subsection{Baselines}
\input{sections/appendix/implementation/baselines.tex}

\section{Experimental Setup}
\input{sections/appendix/dataset}

\section{Extended Examples}
\label{sec:appendix:extended_examples}
\subsection{Theory Examples}
\input{sections/appendix/examples/rebuttal.tex}

\subsection{More Hyper-Elastic Material Models}
\input{sections/appendix/examples/materials.tex}

\section{Incremental Combination of Knowledge for a Newtonian Physics Priors}
\label{sec:appendix:newtonian_priors}
\input{sections/appendix/newtonian.tex}

\section{Complexity Analysis}
\label{sec:complexity}
\input{sections/appendix/complexity.tex}

\end{document}

%% file: tex/dependencies.tex
\usepackage{hyperref}
\usepackage{subcaption}

\usepackage{amsmath}
\usepackage{amssymb}
\usepackage{mathtools}
\usepackage{amsthm}

\usepackage[capitalize,noabbrev]{cleveref}

\usepackage{times}
\usepackage{soul}
\usepackage{url}
\usepackage{graphicx}
\usepackage{amsmath}
\usepackage{amsthm}
\usepackage{amssymb}
\usepackage{booktabs}
\usepackage{algorithm}
\usepackage{algorithmic}
\usepackage{lipsum}
\usepackage{multirow}
\usepackage{multicol}
\usepackage{makecell}
\usepackage{arydshln}
\usepackage[inline]{enumitem}
\usepackage{glossaries}
\usepackage{siunitx}
\usepackage{arydshln}
\usepackage[]{siunitx}

\usepackage{tikz}
\usetikzlibrary{positioning}
\usetikzlibrary{bayesnet}
\usetikzlibrary {arrows.meta}

\usepackage[most]{tcolorbox}
\definecolor{acblue}{rgb}{0.0, 0.45, 0.73}
\newtcbox{\badge}[1][red]{
  on line, 
  arc=2pt,
  colback=#1!50!black,
  colframe=#1!50!black,
  fontupper=\color{white},
  boxrule=1pt, 
  boxsep=0pt,
  left=6pt,
  right=6pt,
  top=2pt,
  bottom=2pt
}

\definecolor{blue}{rgb}{0.0, 0.45, 0.73}
\newtcbox{\tnode}{
  on line, 
  arc=4pt,
  colback=blue!10!white,
  colframe=blue!20!white,
  fontupper=\color{black},
  boxrule=1pt, 
  boxsep=0pt,
  left=2pt,
  right=2pt,
  top=2pt,
  bottom=2pt
}

\theoremstyle{plain}
\newtheorem{theorem}{Theorem}[section]

\theoremstyle{definition}
\newtheorem{definition}[theorem]{Definition}

\theoremstyle{remark}

\newtheorem{example}{Example}

%% file: sections/glossary.tex
\newacronym{AI}{AI}{Artificial Intelligence}
\newacronym{SR}{SR}{Symbolic Regression}
\newacronym{SE}{SE}{Symbolic Expression}
\newacronym{RE}{RE}{Regular Expressions}
\newacronym{RTE}{RTE}{Regular Tree Expression}
\newacronym[plural=pRTEs, firstplural=Probabilistic Regular Tree Expressions (pRTE)]{pRTE}{pRTE}{Probabilistic Regular Tree Expression}
\newacronym{PTA}{PTA}{Probabilistic Tree Automaton}
\newacronym{MCMC}{MCMC}{Markov Chain Monte Carlo}
\newacronym{RJMCMC}{RJ-MCMC}{Reversible Jump MCMC}
\newacronym{HMC}{HMC}{Hamiltonian Monte Carlo}
\newacronym{NUTS}{NUTS}{No U-Turn Sampler}
\newacronym{BSR}{BSR}{Bayesian Symbolic Regression}
\newacronym[plural=CFG, firstplural=Context-Free Grammars (CFG)]{CFG}{CFG}{Context-Free Grammar}
\newacronym{PCFG}{PCFG}{Probabilistic Context-Free Grammar}
\newacronym{NLL}{NLL}{Negative Log-Likelihood}
\newacronym{RMSE}{RMSE}{Root Mean Squared Error}
\newacronym{GP}{GP}{Genetic Programming}

\glsdisablehyper
\makeglossaries

%% file: sections/abstract.tex
\begin{abstract}
\gls{SR} allows for the discovery of scientific equations from data. 
To limit the large search space of possible equations, prior knowledge has been expressed in terms of formal grammars that characterize subsets of arbitrary strings.
However, it is very challenging to formalize the aggregation of two or more items of prior knowledge, because context-free grammars are not closed under intersection. 
To address this issue, we propose to
\begin{enumerate*}[label={(\roman*)}]
    \item compactly express experts' beliefs about which equations are more likely to be expected by \glspl{pRTE}, and
    \item efficiently encode such priors as finite state machines during Bayesian inference for symbolic regression.
\end{enumerate*}
Our experiments show its effectiveness in scientific applications including soil and material science.
\end{abstract}

%% file: sections/introduction.tex
\section{Introduction}

A significant aspect of the scientific process in natural science is to formulate new research hypotheses based on experimental data \citep{schmidt2009distilling}.
\gls{AI} has been proposed as a \textit{new tool} \cite{zdeborova2017new} for scientists to gain understanding \cite{krenn2022scientific} and support hypotheses formulation.
Bayesian methods \cite{jin2019bayesian,guimera2020bayesian} for the long-established task of symbolic regression \cite{cramer1985representation} have become key candidates to aid scientists, because of their well-defined posterior of symbolic solutions given priors and evidence in the form of experimental data.

Centuries of research in natural sciences have yielded a vast amount of knowledge.
Bayesian frameworks can encode such knowledge as prior distributions 
and thus exploit this knowledge to guide the search for symbolic expressions.
Standard approaches to encoding prior knowledge about the structure of the unknown equation would leverage formal languages over \textit{strings}.
These approaches exhibit several weaknesses:
\begin{enumerate*}[label={(\roman*)}]
    \item Sets of syntactically correct arithmetic expressions cannot be described by \textit{regular} languages. The simple language over strings matching the numbers and structures of opening and closing parenthesis, known as \textit{Dyck's language}, is not regular. This renders regular string languages not expressive enough for many languages of practical interest. Thus typically \gls{PCFG} are used. But
    \item \textit{context-free} grammars are not closed under boolean operations\cite{hopcroft2001introduction}, making arbitrary combinations of prior knowledge from different sources infeasible.
\end{enumerate*}
Syntactically correct arithmetic expressions are trees. For example, $\left((x -7)^2 + (x + z)^2 \right) + 12$ is a tree with \tnode{$+$} as the root node, \tnode{$12$} as its child to the right and a continuing subtree to the left.

\begin{figure}[h]
    \centering
    \input{figures/overview}
    \caption{Scientists typically have data and prior knowledge. Our \textit{Bayesian Inference} (\cref{sec:method:bayesian_inference}) requires \textit{samples} (\cref{sec:method:prior_sampling}) and \textit{density evaluations} (\cref{sec:method:prior_evaluation}) of the latter and yields a posterior distribution over expressions that fit the data consistently to the prior knowledge.}
    \label{fig:overview}
\end{figure}

We propose to deal with these drawbacks of \emph{string grammars} by encoding priors using formal \emph{tree languages} \cite{comon2008tree} and suggest a Bayesian inference method for symbolic regression with it. Formal tree grammars can easily be used to \emph{generate trees} and from these trees generate strings, but formal tree grammars cannot be used directly for parsing strings, only for parsing trees. For the previous equation, this means that we can not check whether it is syntactically correct, but given it is we can use a finite state machine that assigns it a probability under a prior.
With these ideas, it is possible to define \textit{expressive} regular languages with:
\begin{enumerate*}[label={(\roman*)}]
    \item compact \gls{pRTE} \cite{weidner2015probabilistic}, and
    \item closure properties \cite{rozenberg2012handbook} under Boolean operations.
\end{enumerate*}

\paragraph{Overview}
For discovering the law of gravitational force a scientist might know that it must be translational invariant.
As depicted in \Cref{fig:overview}, she expresses this \emph{prior knowledge} through a \gls{pRTE}. 
Knowing additionally that the physical units in a force function must match and expressing this through another \gls{pRTE}, \textit{the closure properties} allow us to automatically translate it into a joint \gls{PTA}. 
Our inference algorithm (\cref{sec:method:bayesian_inference}) works on the data interacting with the \gls{pRTE} to generate proposals (\cref{sec:method:prior_sampling}), as well as with the \gls{PTA} to evaluate densities (\cref{sec:method:prior_evaluation}). It yields a \textit{posterior distribution} over arithmetic expressions fitting the data evidence and respecting the experts' prior knowledge.


%% file: figures/overview.tex
\begin{tikzpicture}

\node[draw=gray, rounded corners, align=center] (solver) {Bayesian\\ Inference\\ {\footnotesize (\cref{sec:method:bayesian_inference})}};

\node[left= 1cm of solver] (data) {\includegraphics[width=1.5cm]{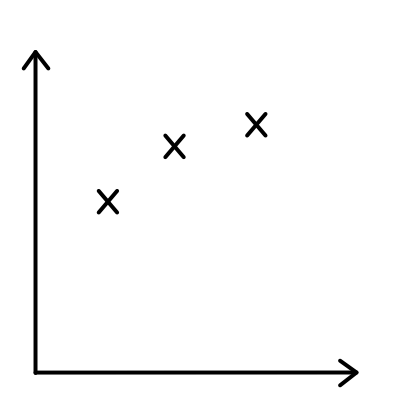}};
\node[below= -2mm of data] (data_label) {Data};
\node[below= 5mm of data_label, xshift=5mm] (knowledge) {};
\node[below= 3mm of data_label, align=center] (knowledge_label) {Experts'\\ Prior \\ Knowledge \\ {\footnotesize (\cref{sec:prior_knowledge})}};

\node[draw= gray, rounded corners, dashed, below= 4mm of solver, xshift=-2em, align=center] (prte) {
pRTE\\
\vspace{-1mm} \\
\vspace{2mm} \scalebox{0.75}{$\left( \tnode{f}(x) \right)^{\infty x}$} \\
{\footnotesize (\cref{sec:method:prior_sampling})}};
\node[draw= gray, rounded corners, dashed, below= 4mm of solver, xshift=15mm, align=center] (wta) {PTA\\
\scalebox{0.4}{
    \input{figures/density_evaluation/reduced_pta.tex}
}\\

{\footnotesize (\cref{sec:method:prior_evaluation})}};

\node[right= 1cm of solver] (posterior) {\includegraphics[width=1.5cm]{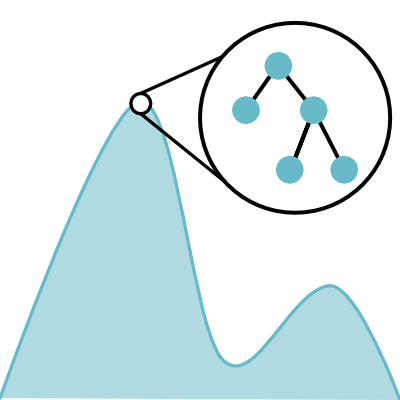}};
\node[below= 0mm of posterior] (posterior_label) {Posterior};


\draw[->, thick] (data) -- (solver);
\draw[->, thick] (solver) -- (posterior);
\draw[->, thick] (knowledge.east |- prte.west) -- (prte.west);
\draw[->, gray] (prte) -- (solver);
\draw[->, gray] (solver) - - (wta);
\draw[->, gray] (prte) - - (wta);

\end{tikzpicture}

%% file: sections/background.tex
\section{Background}
\label{sec:background}

\textit{\gls{SR}} identifies a \textit{symbolic expression}, i.e.\ an analytical equation $f$, that explains a given set of data $\mathcal{D} = \left\{ (x_i,y_i) \right\}_{i=0}^{n}$, i.e.\ $\forall x,y \in \mathcal{D}: f(x) \approx y$. 
There are infinitely many possible solutions $f$ that fit the data up to an error bound, but many of them constitute \emph{physically unreasonable} solutions and thus further preferences may be expressed by \emph{formal languages}.

\subsection{Formal String Languages}
A \textit{formal language} is any subset $L \subseteq \Sigma^*$ of the set of strings $\Sigma^*$ that can be composed out of a finite set of symbols $\Sigma$, called an alphabet.
All finite words $w \in L$ contained in a language can be generated with a set of rules collected in a \textit{grammar}, while the membership of $\Tilde{w} \in \Sigma^*$ in $L$ can be determined through a \textit{automaton}.
Different classes of grammars and automata form a hierarchy of increasing computational complexity of languages, known as \textit{Chomsky Hierarchy} \cite{chomsky1959certain}. 
The simplest class comprises \textit{regular languages}. 
While those have useful properties, e.g. they are efficiently parseable and are closed under Boolean operations, they are not very expressive. Even correct nesting of parenthesis, which is core to arithmetic formulas, cannot be described at an arbitrary depth.
\glspl{CFG} allow for expressing arithmetic formulas at arbitrary depth, but their disadvantage is that they are not closed under intersection. 
Both regular and context-free grammars can be described by production rules that also include probabilities, e.g.\ \gls{PCFG}, and thus express which words belong to a language $L$ and how probable it is that a word $w$ is produced by $L$, i.e. $P(w|L)$.
Let $x,y \in \mathbb{N}^*$ be strings of natural numbers, then $x_\smile y$ is their concatenation.

\subsection{Formal Tree Languages \& Series}
\label{sec:background:tree_languages}

Syntactically correct mathematical expressions are trees.
To characterize sets of trees, one extends the above notion to a \textit{ranked alphabet} $\Sigma = \bigcup_{k \in \mathbb{N}} \Sigma^{(k)}$ giving every symbol $f \in \Sigma^{(k)}$ a $\operatorname{rank}(f) = k$ and thus a defined number of descendants, e.g.\ $f(\cdot, \cdot) \in \Sigma^{(2)}$ and $\operatorname{rank}(f) = 2$. The set of all possible trees over $\Sigma$ is $T_\Sigma$ and every subset $L \subseteq T_\Sigma$ is called a \textit{formal tree language} \cite{comon2008tree}. 

\begin{definition}[Tree]
Let $\Sigma$ be a ranked alphabet, and $D \subseteq \mathbb{N}^*$.
A map $t: D \rightarrow \Sigma$ assigning each node address a symbol in $\Sigma$ is a \textit{tree},
iff it is
\begin{enumerate*}[label={(\roman*)}]
    \item finite, $0 < |D| < \infty$, 
    \item prefix-closed, $u_\smile v \in D \implies u \in D$, and
    \item $\forall x \in D, f = t(x): \{ j | x_\smile j \in D \} = \{0, \ldots, \operatorname{rank}(f) \}\subset \mathbb{N}$.
\end{enumerate*}
\end{definition}
Language membership can be equivalently described by a characteristic function $p: T_\Sigma \rightarrow \{0, 1\}$, called \textit{tree series}, with $p(t) = 1 \iff t \in L$.
With the extension $p: T_\Sigma \rightarrow [0,1]$, the membership is probabilistic and defines a probability distribution over $T_\Sigma$. 
One defines different classes of tree languages. \textit{Regular tree languages} are closed under Boolean operations:
Let $L_1, L_2 \subseteq T_\Sigma$ be regular, then the union $L_1 \cup L_2$, the intersection $L_1 \cap L_2$, the set difference $L_1 \setminus L_2$, and the complement $T_\Sigma \setminus L_1$ are also regular tree languages \cite{rozenberg2012handbook}.
A finite automaton can be constructed to evaluate whether a tree $w\in T_\Sigma$ belongs to a tree language $L$  or not.

\subsubsection{Probabilistic Tree Automata}
\label{sec:background:wta}
 A (top-down) \gls{PTA} \cite{weidner2015probabilistic} is a tuple $(Q, \delta, F, \mu)$ composed of
\begin{enumerate*}[label={(\roman*)}]
    \item a finite set of states $Q$,
    \item a set of final state and symbol pairs $F \subseteq Q \times \Sigma^{(0)}$,
    \item an initial distribution of states $\mu \in \Delta(Q)$, and
    \item a transition function
\end{enumerate*}
\begin{align}
    \delta = \bigcup_n \delta_n \;\;,\; \delta_n: \Sigma^{(n)} \times Q \rightarrow \Delta_0(Q^n),
\end{align}
with $\mathbf{q} \in Q^n$ denoting a tuple of $n$ states. With $\Delta(Q)$ we denote the set of discrete probability distributions over the finite set $Q$. And $\Delta_0 = \Delta \cup \mathbf{0}$ is augmented by the improper distribution $\mathbf{0}$ that assigns zero probability to all states.
Transitions of the automaton $A$ are probabilistic and a successful \textit{run} $r: D \rightarrow Q$ is assigning a state $ q \in Q$ to each node in the tree $t$ such that the states at the leaf nodes are contained in the final set $F$. 

Let $\operatorname{inner}(t), \operatorname{leaf}(t)$ denote the set of inner and respectively leaf nodes of a tree $t$. 
Then the evaluation of the automaton $A$, and thus the value of $p(t)$, is the expectation
\begin{equation}
    \begin{aligned}
        &\mathbb{E}_r \left[ (r(x), t(x)) \in F \;\; \forall \; x \in \operatorname{leaf(t)} \right] =
        &\hspace{-2em} \underset{\substack{\{ r\\ | (r(x), t(x)) \in F \\ \forall \; x \in \operatorname{leaf(t)} \}}}{\sum} \hspace{-1.5em} \mu(r(\epsilon)) \hspace{-1em} \prod_{x \in \operatorname{inner}(t)} \hspace{-1em} \delta_{(t(x), r(x))}\big( r(x_\smile 1), \ldots, r(x_\smile n) \big)
    \end{aligned}
\end{equation}
of runs being successful. Every regular tree series can be recognized by such an automaton \citep{rozenberg2012handbook,weidner2015probabilistic}.

\subsubsection{Probabilistic Regular Tree Expression}
\label{sec:background:prte}
\textit{\gls{pRTE}} \cite{weidner2015probabilistic} compactly assign probabilities to sets of trees.
\begin{example}[\gls{pRTE}]
\label{ex:prte}
\textbf{All trees with at least one \tnode{g}\footnote{To avoid confusion with mathematical symbols, e.g.\ $+$, being elements of a ranked alphabet, we denote $f \in \Sigma$ as $\tnode{f}$.}:}
\begin{equation*}
    \begin{aligned}
        E_1 &= \left(\frac{1}{3}\tnode{f}(x,y) + \frac{1}{3} \tnode{f}(y, x) + \frac{1}{3} \tnode{g}(x) \right)^{\infty y}
        &\circ_x \left( \frac{1}{4} \tnode{f}(x,x) + \frac{1}{4} \tnode{g}(x) + \frac{1}{4}\tnode{a} + \frac{1}{4} \tnode{b} \right)^{\infty x}
    \end{aligned}
\end{equation*}
\end{example}
In \cref{ex:prte}, we denote the set of trees that contain at least one $g$, with three operators:
\begin{enumerate*}[label={(\roman*)}]
    \item \textit{weighted choice} (``$+$") chooses one of the sub-expressions according to the assigned weights,
    \item \textit{concatenation} (``$\circ_x$") replaces each occurrence of the variable $x$ with an expression from the right, and
    \item \textit{infinity iteration} (``$\infty x$") iteratively replaces all occurrences of $x$ (and respectively $y$ for $\infty y$). 
\end{enumerate*}
That way, $E_1$ assigns e.g.\ $g(g(a))$ a probability of $\frac{1}{48} = \frac{1}{3} \cdot \frac{1}{4} \cdot \frac{1}{4}$ and $f(b,b)$ a probability of $0$. 
An extended \cref{ex:extended_prte_example} that explains this process step by step is provided in \cref{sec:appendix:extended_examples}.


%% file: sections/related.tex
\section{Related Work}
\label{sec:related_work}

\paragraph{Symbolic Regression}
has been approached from various perspectives:
\begin{enumerate*}[label={(\roman*)}]
    \item \textit{Informed Search Strategies} ranging from classical specialized evolution strategies in genetic programming \cite{koza1990genetic,cramer1985representation} to select the best expression among a population to guiding search \cite{udrescu2020ai,udrescu2020ai2,landajuela2022unified} by \textit{heuristics} inspired from science like symmetry and separability tests;
    \item \textit{Optimization Based} by first solving a continuous optimization like \textit{LASSO} in a library of terms \cite{brunton2016discovering} or optimizing a neural network \cite{sahoo2018learning} and extracting a symbolic expression from it afterwards;
    \item \textit{Representation Learning} \cite{kusner2017grammar,mevznar2023efficient} to find an embedding space for equations that eases the regression task for classical search,
    \item \textit{Deep Learning Approaches} \cite{petersen2019deep,valipour2021symbolicgpt,kamienny2022end,holt2023deep} train neural networks supervised or with a reinforcing objective to predict matching expressions, and
    \item \textit{Bayesian Approaches}, as described next.
\end{enumerate*}

\paragraph{Bayesian Approaches}
need to combine reasoning about the symbolic structure of an equation with reasoning about (continuous) parameters inside it.
Proposing changes to the tree structure of an equation with a Markov Chain \cite{jin2019bayesian,guimera2020bayesian}
and deal with varying aomounts of the parameters among symbolic candidates by
\begin{enumerate*}[label={(\roman*)}]
    \item reversible jumps \cite{jin2019bayesian} in a joint space of parameters and auxiliary variables, or
    \item selecting structure candidates based on the \textit{Bayesian Information Criterion (BIC)} \cite{guimera2020bayesian}. 
\end{enumerate*}
While the latter is only a first-order approximation to the true posterior \cite{giraud2021introduction} and cuts the search space at a predefined depth, the former hard-codes a prior that does not generalize well in benchmarks \cite{la2021contemporary}.
Assigning probabilities to \textit{dictionaries} that capture discrete and continuous choices taken in a \emph{probabilistic program} \cite{cusumano2020automating} has symbolic regression as a special case. 
All these approaches lack a compact and easily exchangeable representation of scientific prior knowledge.

\paragraph{Scientific Prior Knowledge} has mostly been expressed as context-free grammars over strings.
This includes general or domain-specific assumptions about equations in Backus Naur Form (BNF) \cite{lu2016using}, PCFG \cite{brence2021probabilistic,xu2021bayesian} and attribute grammars \cite{brence2023dimensionally}. Likewise, the search strategy in \citet{udrescu2020ai} restricting to certain combinations of subterms on detected invariance can be understood as \gls{CFG}. To incrementally improve predictions and knowledge \gls{CFG} have been used for interactive feedback \cite{poli1997genetic,crochepierre2022interactive}.
Beyond grammars, random graph models \cite{guimera2020bayesian} have bin used to learn the inherent structure of equations on \textit{Wikipedia}.

%% file: sections/method.tex
\section{Symbolic Reasoning with Probabilistic Regular Tree Priors}
\label{sec:method}

\begin{figure*}[t]
     \centering
     \begin{subfigure}[b]{0.25\textwidth}
         \centering
         \scalebox{0.7}{\input{figures/density_evaluation/symbolic_example}}
         \caption{$(x -7)^2 + (x + z)^2 + 12$}
         \label{fig:symbolic_example}
     \end{subfigure}
     \hfill
     \begin{subfigure}[b]{0.25\textwidth}
         \centering
         \scalebox{0.7}{\input{figures/density_evaluation/structure_tree}}
         \caption{$t \in T_\Sigma$}
         \label{fig:symbolic_example2}
     \end{subfigure}
     \hfill
     \begin{subfigure}[b]{0.4\textwidth}
         \centering
         \scalebox{0.7}{\input{figures/density_evaluation/factor_graph}}
         \caption{factor graph with observations $x_i$ and states $q_i$}
         \label{fig:factor_graph}
     \end{subfigure}
    \caption{
    We represent an expression (\subref{fig:symbolic_example}) in variables $x,y,z$ with a tree $t\in T_\Sigma$ (\subref{fig:symbolic_example2}) and $c \in \Sigma^{(0)}$ marking the positions of parameters $\theta_c = (7, 12)^T$. 
    To evaluate a regular tree prior $p(t)$ we construct a factor graph (\subref{fig:factor_graph}):
    For every node in $t$, there are \textit{two} random variables, $x_i$ and $q_i$.
    The random variables $x_i$ model the possible ranked symbols for a tree node and are thus observed, e.g.\ $x_5$ corresponds to $\tnode{-} \in \Sigma^{(0)}$ in $t$. 
    and assignments to the random variable $q_5$ model the possible states a \gls{PTA} can be in after parsing the tree $t$ up to that symbol. Finally, factors $\phi_j$ express the transition probabilities of the \gls{PTA}. 
    To determine $p(t)$ we condition on all $x_i$ (i.e.\ the symbols in the input tree $t$) and marginalize over all possible state assignments to $q_i$ (i.e.\ runs of the \gls{PTA}).}
    \label{fig:factor_graph_construction}
\end{figure*}

We propose to model the solution space (\cref{sec:method:sr_definition}) for symbolic regression in terms of tree languages 
together with probability distributions over that space. With that, the task of symbolic regression can be naturally expressed as a Bayesian inference step (\cref{sec:method:bayesian_inference}), 
and a multitude of prior knowledge (compare \cref{sec:application:sorption}) can be expressed through probabilistic regular tree expressions. 

\subsection{Symbolic Expressions as Tree Languages }
\label{sec:method:sr_definition}

First, we divide symbolic expressions into their syntactic structure and parameters.
\begin{definition}[Symbolic Expression]
\label{def:symbolic_expression}
Let $\Sigma$ be ranked alphabet with special symbols $c,d\in\Sigma^{(0)}$. A symbolic expression $\mathfrak{e} = (t, \theta_d, \theta_c) \in \mathfrak{E}$ is a tuple of a symbolic tree structure $t \in T_\Sigma$, discrete $\theta_d\in\Theta_d(t)$, and continuous $\theta_c \in \Theta_c(t)$ parameter vectors, s.t.\ 
$\Theta_c(t) := \mathbb{R}^{|\operatorname{pos}_c(t)|}$ and 
$\Theta_d(t) := \mathbb{Z}^{|\operatorname{pos}_d(t)|}$ hold
with $\operatorname{pos}_x = \{ d \in D| t(d) = x \} \; \forall x \in \Sigma$ denoting the set of all positions of a symbol.
\end{definition}
In other words, 
the syntax is modeled through a tree language
and parameters are kept in separate vector spaces $\Theta_d, \Theta_c$ with special symbols $c, d \in \Sigma^{(0)}$ marking their locations in the expression.
$\Theta_d, \Theta_c$ can not directly be part of the ranked alphabet which is required to be \textit{finite}.

\begin{example}
    \label{ex:symbolic_expression_main}
    Let $\Sigma = \Sigma^{(0)} \cup \Sigma^{(1)} \cup \Sigma^{(2)}$ with
    \begin{align*}
        \Sigma^{(0)} &:= \left\{ \tnode{x}, \tnode{c}, \tnode{d} \right\} &
        \Sigma^{(1)} &:= \left\{ \tnode{\tiny $\sin$} \right\} &
        \Sigma^{(2)} &:= \left\{ \tnode{+}, \tnode{$\cdot$}, \tnode{\tiny pow} \right\}.
    \end{align*}
A symbolic expression $\mathfrak{e} = (t, \theta_d, \theta_c) \in \mathfrak{E}$ is defined by $t = \tnode{\footnotesize pow}(\tnode{+}(\tnode{$\cdot$}(\tnode{c},\tnode{$\sin$}(\tnode{x})), \tnode{c}), \tnode{d})$, $\operatorname{pos}_c = \{ 01, 000 \} \subseteq D$, $\operatorname{pos}_d = \{ 1 \} \subseteq D$, $\Theta_c(t) := \mathbb{R}^{2}$, $\Theta_d(t) := \mathbb{Z}$ and e.g.\ $\theta_c = (0.25, 1.06)^T \in \Theta_c(t)$, $\theta_d = 2 \in \mathbb{Z}$. This yields the symbolic expression $(\sin(0.25 x) + 1.06)^2$.
\end{example}

The unrestricted solution space of a symbolic regression task is
\begin{equation}
    \mathfrak{E} := \bigcup_{t \in T_\Sigma} \Big\{ (t, \theta_d, \theta_c) \;\Big| \;\; \theta_d \in \Theta_d(t); \theta_c \in \Theta_c(t) \Big\}.
\end{equation}
We assign probabilities $0 \leq p(\mathfrak{e}) \leq 1$ to expressions $\mathfrak{e} \in \mathfrak{E}$ such that overall
\begin{equation}
    \label{eq:total_probaility_symbolic_expression}
    p(\mathfrak{E}) = \sum_{\mathfrak{e} \in \mathfrak{E}} p(\mathfrak{e}) = \sum_{t \in T_\Sigma} \Bigg( \sum_{\theta_d \in \Theta_d(t)} \Big( \int_{\Theta_c(t)} p(t)p(\theta_c | t)p(\theta_d | t) \;d\theta_c \Big) \Bigg) = 1
\end{equation}
holds. 
This construction, inspired from \cite{cusumano2020automating} for probabilistic programs, splits the measure into a measure $p(t)$ for trees, as well as measures $p(\theta_c | t)$, $p(\theta_d | t)$ for parameters. 
While the latter can be realized through standard Lebesgue and counting measures, $p(t) = \frac{1}{Z} \tilde{p}(t)$ is a \emph{formal tree series} with $Z = \sum_{t \in T_\Sigma} \Tilde{p}(t) < \infty$ inducing a probability for tree structures.
\subsection{Sampling-Based Bayesian Inference}
\label{sec:method:bayesian_inference}

Having defined symbolic expressions $\mathfrak{e}$ in terms of tree languages, the task of symbolic regression given data $\mathcal{D} = (\mathcal{X}, \mathcal{Y})$, is expressed through a Bayesian inference step as
\begin{equation}
    \label{eq:sr_posterior}
    \begin{aligned}
        p(\mathfrak{e} | \mathcal{X}, \mathcal{Y}) &\propto p(\mathcal{X}, \mathcal{Y} | \mathfrak{e}) p(\mathfrak{e})
        \propto p(\mathcal{X}, \mathcal{Y} | t, \theta_c, \theta_d) \Tilde{p}(t) p(\theta_d |t) p(\theta_c | t),
    \end{aligned}
\end{equation}
where we assume the data to be noisy observations $\mathcal{Y} \sim \mathcal{N}(\operatorname{eval}_\mathfrak{e}(\mathcal{X}), \sigma)$ of the evaluation of $\mathfrak{e}$ at $\mathcal{X}$ corrupted by Gaussian noise of some standard deviation $\sigma \sim \operatorname{Exp}(\lambda)$,
whose value is jointly inferred.
Thus the likelihood $p(\mathcal{X}, \mathcal{Y} | t, \theta_c, \theta_d)$ is Gaussian.
Finally, a tree series $\Tilde{p}(t)$ and prior distributions for the parameters $p(\theta_d |t)$, $p(\theta_c | t)$ jointly express scientists' prior knowledge.
We call the combination of these three priors a \textit{regular tree prior} $p(\mathfrak{e})$ and show its usefulness for scientific symbolic regression.

\begin{figure*}[t]
    \centering
    \begin{subfigure}[b]{0.19\textwidth}
        \centering
        \includegraphics*[width=\textwidth]{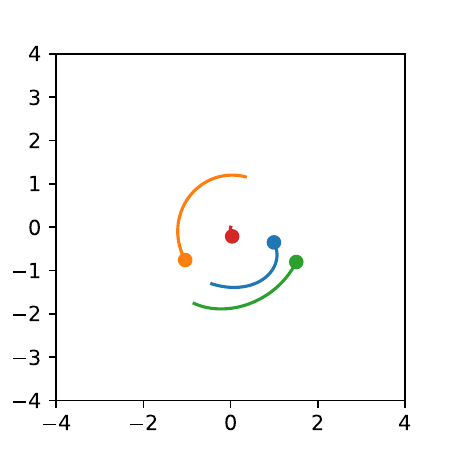}
        \caption{Training Data}
        \label{fig:newtonian_example_data}
    \end{subfigure}
    \hfill
    \begin{subfigure}[b]{0.19\textwidth}
        \centering
        \includegraphics*[width=\textwidth]{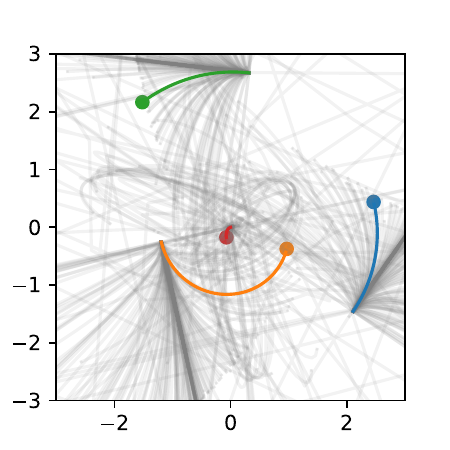}
        \caption{Prior}
        \label{fig:newtonian_example_prior}
    \end{subfigure}
    \hfill
    \begin{subfigure}[b]{0.19\textwidth}
        \centering
        \includegraphics*[width=\textwidth]{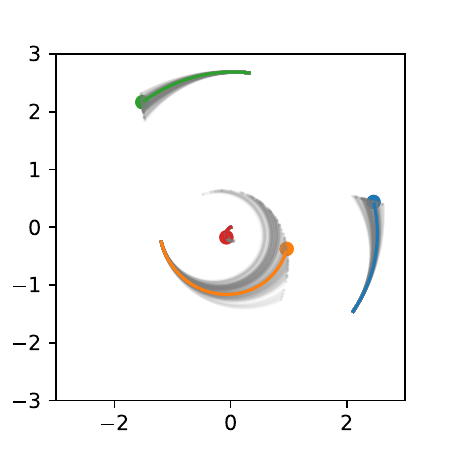}
        \caption{Posterior}
        \label{fig:newtonian_example_posterior}
    \end{subfigure}
    \hfill
    \begin{subfigure}[b]{0.4\textwidth}
        \centering
        \includegraphics*[width=\textwidth]{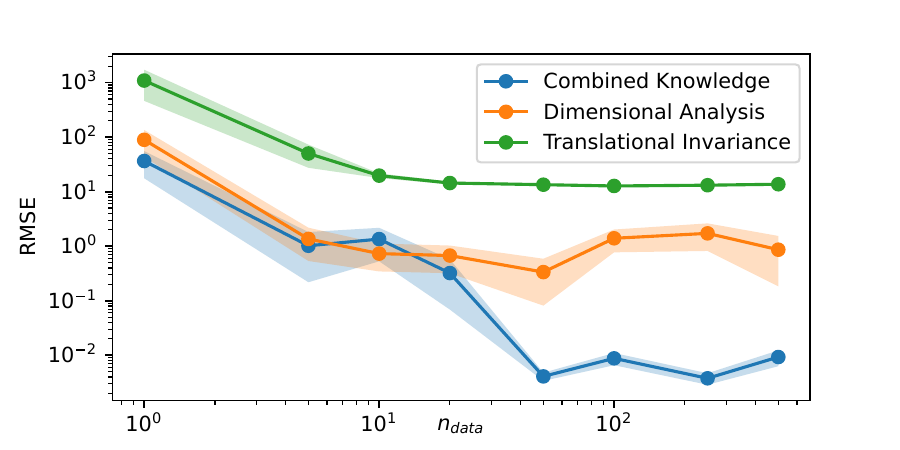}
        \caption{Results}
        \label{fig:newtonian_example_result}
    \end{subfigure}
   \caption{Particle system: system knowledge defines a prior distribution of potential force functions (\subref{fig:newtonian_example_prior}), which results after inference with training data (\subref{fig:newtonian_example_data}) in a posterior distribution (\subref{fig:newtonian_example_posterior}). More data and combinations of knowledge (\subref{fig:newtonian_example_result}) help to find the true gravitational force. In (\subref{fig:newtonian_example_data}, \subref{fig:newtonian_example_prior}, \subref{fig:newtonian_example_posterior}) colored paths show the true trajectories of the particles in the corresponding datasets.}
   \label{fig:newtonian_example}
\end{figure*}

The posterior $p(\mathfrak{e} | \mathcal{X}, \mathcal{Y})$ is approximated through sampling expressions $\mathfrak{e}'$ (given $\mathfrak{e}$) from a Markov Chain with a \emph{Metropolis-Hastings} \cite{metropolis1953equation,hastings1970monte} transition kernel
\begin{equation}
    \label{eq:mcmc_transition_kernel}
    \mathcal{T}(\mathfrak{e} \rightarrow \mathfrak{e}') = \mathcal{T}^Q(\mathfrak{e} \rightarrow \mathfrak{e}')\mathcal{A}(\mathfrak{e} \rightarrow \mathfrak{e}'),
\end{equation}
composed of a proposal $\mathcal{T}^Q(\mathfrak{e} \rightarrow \mathfrak{e}')$ followed by a Hastings step  with acceptance probability
\begin{equation}
    \label{eq:acceptance_prob}
    \begin{aligned}
        \mathcal{A}(\mathfrak{e} \rightarrow \mathfrak{e}') = \min \left[ 1 ,\frac{p(X, Y | t', \theta_c', \theta_d') \tilde{p}(t') p(\theta_c' | t) p(\theta_d' | t)}{p(X, Y | t, \theta_c, \theta_d) \tilde{p}(t) p(\theta_c | t) p(\theta_d | t) }
        \cdot \frac{p(u') \mathcal{T}^Q(\mathfrak{e}' \rightarrow \mathfrak{e})}{p(u) \mathcal{T}^Q(\mathfrak{e} \rightarrow \mathfrak{e}')} \det \Big(\frac{\partial (\theta', u')}{\partial (\theta, u)} \Big) \right],
    \end{aligned}
\end{equation}
that can be determined by evaluations of the tree series $\Tilde{p}$.
Changing the tree structure $t$ to $t'$ may affect the number of parameters contained in the corresponding expressions $\mathfrak{e}, \mathfrak{e}'$. To make these jumps \textit{reversible} \cite{green1995reversible} (i.e.\ $\mathcal{T}^Q(\mathfrak{e}' \rightarrow \mathfrak{e})$ exists) we sample as many auxiliary variables $u, u'$ to match the dimensions $\dim([\theta, u]) = \dim([\theta', u'])$.
Jumps in that joint space $[\theta, u]$ may change the volume corrected by a change of variable formula (\cref{sec:appendix:logdetjacobian}).


\subsection{Sampling Symbolic Expressions}
\label{sec:method:prior_sampling}
Proposals should explore the space of potential symbolic expressions well but also exploit the experts' prior knowledge of promising expressions.
To that end, $\mathcal{T}^Q(\mathfrak{e} \rightarrow \mathfrak{e}')$ splits into transit kernels for the tree structures $\mathcal{T}^Q(t \rightarrow t')$ and for the parameters $\mathcal{T}^Q(\theta \rightarrow \theta')$.
While $\mathcal{T}^Q(\theta_c \rightarrow \theta'_c)$ can exploit gradients and thus leverage efficient Hamiltonian Monte Carlo samplers,
$\mathcal{T}^Q(t \rightarrow t')$ proposes discrete changes without gradient information.
These discrete proposals could be
\begin{enumerate*}[label={(\roman*)}]
    \item \textit{global moves} by sampling (a variant of) the prior,
    \item \textit{local moves} by re-sampling contexts in the grammar (\cref{sec:appendix:context_resampling}), or
    \item a neural network's prediction.
\end{enumerate*}
We limit our experiments to the first option and leave other options as a promising direction for better exploration of structures in future work. 
Once a new tree structure is determined, all occurrences $\tnode{c}, \tnode{d}$ are replaced by samples $c \in \mathbb{R}, d \in \mathbb{Q}$ from the priors $p(\theta_c | t), p(\theta_d | t)$ or a kernel $\mathcal{T}^Q(\theta_c \rightarrow \theta'_c)$ respectively. 

\subsection{Regular Tree Prior Density Evaluation}
\label{sec:method:prior_evaluation}

To evaluate $p(\mathfrak{e})$ and respectively $\tilde{p}(t)$, e.g., for the acceptance probability (\cref{eq:acceptance_prob}) of a proposal, we construct a \textit{factor graph} \cite{koller2009probabilistic} that resembles the structure of the tree $t$ to be evaluated as shown in \Cref{fig:factor_graph_construction}.
Recall that a run of a \gls{PTA} is an assignment of states given a tree $t$ as an input.
Each node in $t$ corresponds to a pair of \emph{random variables} $q_i, x_i$ in the factor graph. The latent variable $q_i$ models the state assignment in a run of the \gls{PTA} (derived from $\Tilde{p}$; compare \cref{sec:background:wta}) and the observed variable $x_i$ models the input of a \textit{ranked symbol} in position $i \in D \subseteq \mathbb{N}^*$, i.e.\ $t(i)$. 
A factor $\phi_j \in V_1 \times \ldots \times V_{n+1}$ in the graph is a tensor of rank $n+1$ with $V_k = \mathbb{R}_+^{|Q|} \;\; \forall k$  and corresponds to an entry of the transition function $\delta$ of the \gls{PTA} for a symbol of rank $n$, as illustrated in the following example.
\begin{example}[Factor Graph]
    \label{example:factor_graph}
    Consider a \gls{PTA} has read the symbol $\tnode{+} \in \Sigma$ at the root of a tree $t$ in \Cref{fig:factor_graph_construction} and is about to read the next symbol in the left subtree.
    In \Cref{fig:factor_graph_construction}\subref{fig:factor_graph}, the random variable $x_1$ models this next symbol to be read and the variable $q_1$ models the state of the \gls{PTA} before doing so.
    The factor $\phi_1$ assigns a probability for the \gls{PTA} to start in a state assignment to $q_1$, reading $x_1$ to be again $\tnode{+} \in \Sigma$, and \emph{transition} to individual state assignments for both, $q_3$ and $q_4$. The factor $\phi_0$ models the initial state distribution $\mu$.
\end{example}
Since $\tilde{p}(t)$ is the expectation of runs being successful, it corresponds to $p(x_1, \ldots, x_n) = \sum_{q_1, \ldots, q_m} \prod_l \phi_l$ marginalizing all possible state assignments.


%% file: figures/density_evaluation/symbolic_example.tex
\definecolor{blue}{rgb}{0.0, 0.45, 0.73}
\begin{tikzpicture}

\tikzstyle symbolic node=[draw, blue, thick, fill=blue!10!white, circle, minimum size=2em, text=black];
\tikzstyle symbolic link=[draw, blue, very thick];

\node[symbolic node] (plus1) {$+$};
\node[symbolic node, below= 0.5em of plus1, xshift= -1.5em] (plus2) {$+$};
\node[symbolic node, below= 0.5em of plus1, xshift= 5em] (twelve) {$12$};

\node[symbolic node, below= 0.5em of plus2, xshift= -3em] (sq1) {$\cdot^{2}$};
\node[symbolic node, below= 0.5em of plus2, xshift= 3em] (sq2) {$\cdot^{2}$};

\node[symbolic node, below= 0.5em of sq1] (minus) {$-$};
\node[symbolic node, below= 0.5em of minus, xshift=-1.5em] (x) {$x$};
\node[symbolic node, below= 0.5em of minus, xshift=1.5em] (seven) {$7$};
\node[symbolic node, below= 0.5em of sq2] (plus3) {$+$};
\node[symbolic node, below= 0.5em of plus3, xshift=-1.5em] (y) {$y$};
\node[symbolic node, below= 0.5em of plus3, xshift=1.5em] (z) {$z$};

\draw[symbolic link] (plus1) -- (plus2);
\draw[symbolic link] (plus1) -- (twelve);
\draw[symbolic link] (plus2) -- (sq1);
\draw[symbolic link] (plus2) -- (sq2);
\draw[symbolic link] (sq1) -- (minus);
\draw[symbolic link] (sq2) -- (plus3);
\draw[symbolic link] (x) -- (minus);
\draw[symbolic link] (seven) -- (minus);
\draw[symbolic link] (y) -- (plus3);
\draw[symbolic link] (z) -- (plus3);

\end{tikzpicture}

%% file: figures/density_evaluation/structure_tree.tex
\definecolor{blue}{rgb}{0.0, 0.45, 0.73}
\begin{tikzpicture}

\tikzstyle symbolic node=[draw, blue, thick, fill=blue!10!white, circle, minimum size=2em, text=black];
\tikzstyle symbolic link=[draw, blue, very thick];

\node[symbolic node] (plus1) {$+$};
\node[symbolic node, below= 0.5em of plus1, xshift= -1.5em] (plus2) {$+$};
\node[symbolic node, below= 0.5em of plus1, xshift= 5em] (twelve) {$c$};

\node[symbolic node, below= 0.5em of plus2, xshift= -3em] (sq1) {$\cdot^{2}$};
\node[symbolic node, below= 0.5em of plus2, xshift= 3em] (sq2) {$\cdot^{2}$};

\node[symbolic node, below= 0.5em of sq1] (minus) {$-$};
\node[symbolic node, below= 0.5em of minus, xshift=-1.5em] (x) {$x$};
\node[symbolic node, below= 0.5em of minus, xshift=1.5em] (seven) {$c$};
\node[symbolic node, below= 0.5em of sq2] (plus3) {$+$};
\node[symbolic node, below= 0.5em of plus3, xshift=-1.5em] (y) {$y$};
\node[symbolic node, below= 0.5em of plus3, xshift=1.5em] (z) {$z$};

\draw[symbolic link] (plus1) -- (plus2);
\draw[symbolic link] (plus1) -- (twelve);
\draw[symbolic link] (plus2) -- (sq1);
\draw[symbolic link] (plus2) -- (sq2);
\draw[symbolic link] (sq1) -- (minus);
\draw[symbolic link] (sq2) -- (plus3);
\draw[symbolic link] (x) -- (minus);
\draw[symbolic link] (seven) -- (minus);
\draw[symbolic link] (y) -- (plus3);
\draw[symbolic link] (z) -- (plus3);

\end{tikzpicture}

%% file: figures/density_evaluation/factor_graph.tex
\begin{tikzpicture}
    \node[latent] (q0) {$q_0$};
    \node[latent, below= 0.5em of q0, xshift= -1.5em] (q1) {$q_1$};
    \node[latent, below= 0.5em of q0, xshift= 3em] (q2) {$q_2$};
    \node[latent, below= 0.5em of q1, xshift= -5em] (q3) {$q_3$};
    \node[latent, below= 0.5em of q1, xshift= 5em] (q4) {$q_4$};
    \node[latent, below= 0.5em of q3] (q5) {$q_5$};
    \node[latent, below= 0.5em of q4] (q6) {$q_6$};
    \node[latent, below= 0.5em of q5, xshift= -1.5em] (q7) {$q_7$};
    \node[latent, below= 0.5em of q5, xshift= 1.5em] (q8) {$q_8$};
    \node[latent, below= 0.5em of q6, xshift= -1.5em] (q9) {$q_9$};
    \node[latent, below= 0.5em of q6, xshift= 1.5em] (q10) {$q_{10}$};
    
    \node[obs, right= 1em of q0] (x0) {$x_0$};
    \node[obs, left= 1em of q1] (x1) {$x_1$};
    \node[obs, right= 1em of q2] (x2) {$x_2$};
    \node[obs, left= 1em of q3] (x3) {$x_3$};
    \node[obs, right= 1em of q4] (x4) {$x_4$};
    \node[obs, left= 1em of q5] (x5) {$x_5$};
    \node[obs, right= 1em of q6] (x6) {$x_6$};
    \node[obs, left= 1em of q7] (x7) {$x_7$};
    \node[obs, right= 1em of q10] (x10) {$x_{10}$};
    \node[obs, right= 0.25em of q8, yshift=2em] (x8) {$x_8$};
    \node[obs, left= 0.25em of q9, yshift=2em] (x9) {$x_9$};
    
    \factor[-, left= 0.5em of q0] {} {left:$\phi_0$} {q0} {};
    \factor[-, below= 0.25em of q0] {} {} {x0, q0, q1, q2} {};
    \factor[-, below= 0.5em of q1] {} {below:$\phi_1$} {x1, q1, q3, q4} {};
    \factor[-, right= 0.25em of q2] {} {} {x2, q2} {};
    \factor[-, below= 0.0em of q3, xshift=-1em] {} {} {x3, q3, q5} {};
    \factor[-, below= 0.0em of q4, xshift=1em] {} {} {x4, q4, q6} {};
    \factor[-, below= 0.25em of q5] {} {} {x5, q5, q7, q8} {};
    \factor[-, below= 0.25em of q6] {} {} {x6, q6, q9, q10} {};
    \factor[-, left= 0.25em of q7] {} {} {x7, q7} {};
    \factor[-, right= 0.25em of q8] {} {} {x8, q8} {};
    \factor[-, left= 0.25em of q9] {} {} {x9, q9} {};
    \factor[-, right= 0.25em of q10] {} {} {x10, q10} {};
\end{tikzpicture}

%% file: sections/boolean.tex
\begin{figure*}[ht]
    \centering
    \begin{subfigure}[b]{0.32\textwidth}
        \centering
        \includegraphics[width=\textwidth]{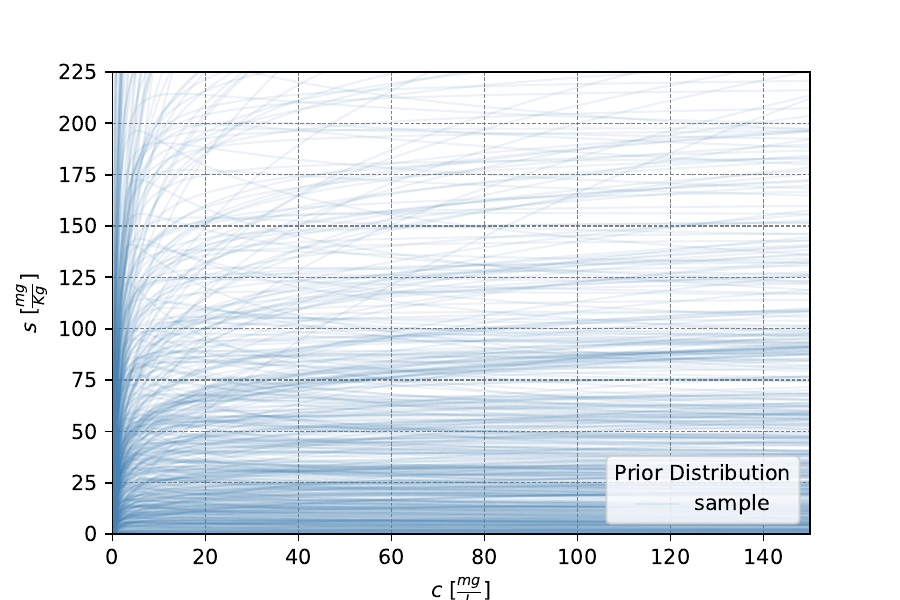}
        \caption{Prior Distribution}
        \label{fig:prte_prior}
    \end{subfigure}
    \hfill
    \begin{subfigure}[b]{0.32\textwidth}
        \centering
        \includegraphics[width=\textwidth]{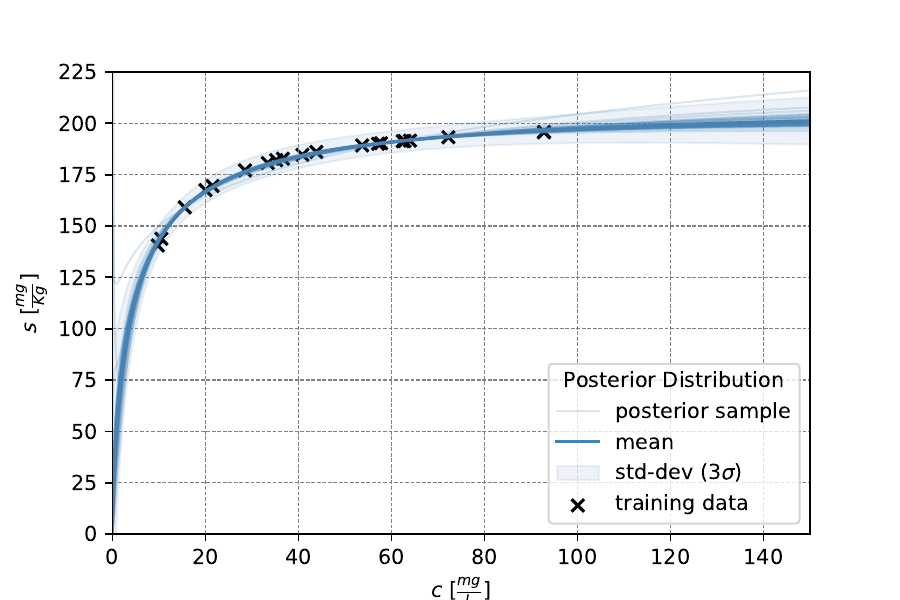}
        \caption{Posterior Distribution}
        \label{fig:prte_posterior}
    \end{subfigure}
    \hfill
    \begin{subfigure}[b]{0.32\textwidth}
        \centering
        \includegraphics[width=\textwidth]{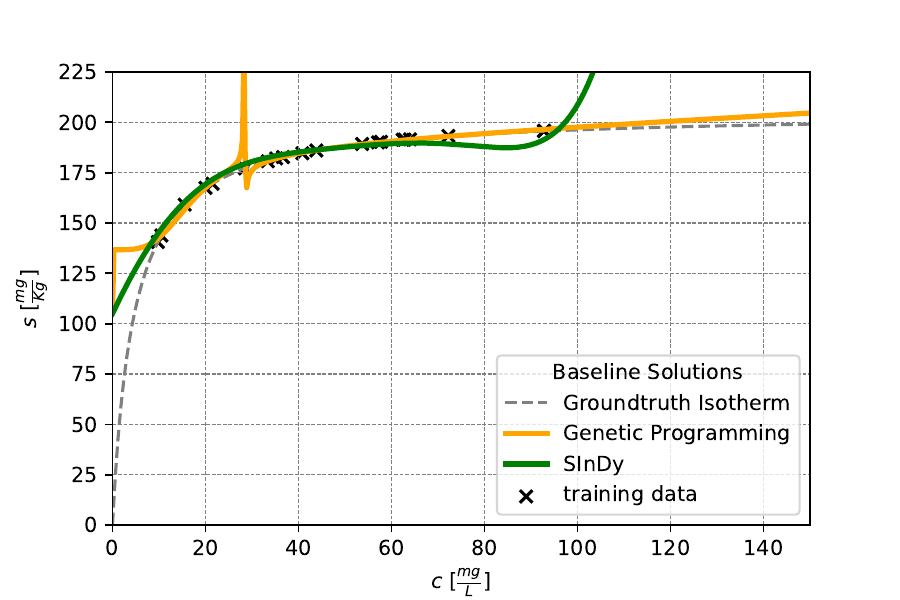}
        \caption{Baseline Solutions}
        \label{fig:baseline_results}
    \end{subfigure}
    \caption{Example of synthetic sorption isotherm (Langmuir) data, with (\subref{fig:prte_prior}) samples from the prior encoded in the regular expression, (\subref{fig:prte_posterior}) the posterior of equations given data evidence. In contrast a typical baseline solution (\subref{fig:baseline_results}) yields unphysical predictions on the same data.}
    \label{fig:visual_results}
\end{figure*}

\section{Prior Knowledge from Boolean Combinations}
\label{sec:prior_knowledge}

Science rarely occurs in a leap. 
A hypothesis is built, rejected and refined many times. 
Thus symbolic regression's capabilities to incorporate prior knowledge must reflect this.
Key to the proposed formalism is its \textit{closure under boolean operations}, which directly transfers from the known closure \cite{rozenberg2012handbook} of regular tree languages:
To find the law of \textit{gravitational force} from observations of particle trajectories, one can exploit two assumptions \cite{xu2021bayesian}.
First, with \textit{translational invariance} one can restrict the solution space to functions wherein positions of particles $p_i, p_j$ only occur as differences. 
Second, with \textit{dimensional analysis}, only quantities with matching units are allowed to be added or subtracted. 
\citet{xu2021bayesian} hand-craft a context-free grammar that conjoins these two aspects. 
We represent the same knowledge with the intersection of two independently specified \glspl{pRTE} (\cref{sec:appendix:newtonian_priors}).

\subsection{Orbiting Particle Systems}
We apply this knowledge to an $N$-body system with a stationary particle of high mass at the center and $N -1$ particles of random but lower mass orbiting around it. 
We follow the setup of \cite{xu2021bayesian} but leave the particle masses $m_i, m_j$ observable. 
Noisy training data $\{(p_i, p_j, m_i, m_j, v_i, v_j, f_{ij} + \epsilon)\}_{ij}$ from the interaction of pairs of particles is derived from the system in \cref{fig:newtonian_example_data}. 
We use three hold-out particle systems for testing.
The error of the learned force functions $f(p_i, p_j, m_i, m_j, v_i, v_j)$ is shown in \cref{fig:newtonian_example_result} to decrease for an increasing amount of training data. 
For the prior only encoding \textit{translational invariance} the resulting function space is too broad to let the inference converge to a good solution within the computational budget. 
Through combination with \textit{dimensional analysis}, the true force law is recovered (\cref{fig:newtonian_example_posterior}) even though the prior (\cref{fig:newtonian_example_prior}) was quite uninformative. 
The remaining uncertainty stems from jointly estimating the gravitation constant.

But context-free grammars are not closed under Boolean operations \cite{hopcroft2001introduction}, prohibiting this incremental refinement of priors in general. With closure properties experts could express: I know it is either this ($E_1$) or less likely this ($E_2$) but not this ($E_3$); i.e.\ $(E_1 \cup E_2) \setminus E_3$.

%% file: sections/application.tex
\section{Scientific Case Study}
\label{sec:application}

\renewcommand\theadfont{\bfseries}
\begin{table*}[t]
    \caption{Regression of Sorption Isotherms. Best performance (avg.\ RMSE) bolded per dataset.}
    \label{tab:syn_isotherm_results}
    \vskip 0.15in
    \begin{center}
    \begin{small}
    \begin{sc}
        \input{tables/sorption/isotherm_minimal.tex}
    \end{sc}
    \end{small}
    \end{center}
    \vskip -0.1in 
\end{table*}

Existing benchmark datasets \cite{la2021contemporary,udrescu2020ai} often compromise a broad variety of equations
for which it is infeasible to come up with general knowledge for fitting functions.
Many scientific applications differ by
\begin{enumerate*}[label={(\roman*)}]
    \item the existence of sophisticated but domain-specific knowledge about which solutions can be considered,
    \item the data sparsity due to expensive acquisition in experiments.
\end{enumerate*}
In this section, we show the scientific discovery process with case studies from soil and material science.

\paragraph{Experimental Setup}
\label{sec:application:setup}
We randomly sample inputs and parameters for models from domain-specific literature (see \cref{sec:synthetic_dataset}) and corrupt the outputs with noise to generate challenging symbolic regression problems.
In order not to give an unfair advantage to our method, we have limited the usefulness of prior knowledge to such prior knowledge as is typically available for scientific problems, such as the structure of equations.
Most importantly, for the parameters $\theta_c, \theta_d$ in the expressions, we have chosen uninformative priors.
The true values for the parameters are on average $6.8466$ standard deviations away from the mean of the prior distribution. 
The ratio of p-values between the prior expectation and actual value looked up in the prior density is $0.0318$.
We compare the performance of our method with a variety of standard symbolic regression from genetic programming \citep{koza1990genetic}, over sparse identification \citep{brunton2016discovering} to recent deep learning approaches \citep{kusner2017grammar,holt2023deep} (details in \cref{sec:appendix:implementation_details}).

\subsection{Sorption Isotherms}
\label{sec:application:sorption}
The transition of ions or molecules from a solution to a solid phase is called \textit{sorption}.  
In literature, there exists a multitude of sorption isotherm equations to relate the \textit{equilibrium sorptive concentration} $c$ ($\frac{\text{mg}}{L}$) in the solution phase and the \textit{sorbate concentration} $s$ ($\frac{\text{mg}}{\text{Kg}}$) in the soil to characterize the retention of contaminants in soils and sediments \cite{thompson2012sorption}.
These models have been unified \cite{hinz2001description} as 
\begin{equation}
    \label{eq:sorption_knowledge}
    s = s_T \sum_{i = 1}^{n} f_i \prod_{j=1}^{m_i} \left( \frac{q_{ij} c^{\alpha_{ij}}}{1 + p_{ij} c^{\beta_{ij}}} \right)^{\gamma_{ij}},
\end{equation}
with each choice for $n, m_0, \ldots, m_n$ and assignment to parameters $s_T, f_i, q_{ij}, p_{ij}, \alpha_{ij}, \beta_{ij}, \gamma_{ij}$ being a model that relates $s$ and $c$.
All of these models share some regular pattern which we express with a regular tree expression $E_{\text{iso}}$ as
\begin{equation}
    \label{eq:sorption_prte}
    \begin{aligned}
        &\tnode{$\cdot$}\left( \tnode{$s_T$}, \left( \rho_s \cdot \tnode{$+$}(y, x) + (1 - \rho_s) \cdot y \right)^{\infty x} \right)
        \circ_{y} \tnode{$\cdot$} \left( \tnode{$f$}, \left( \rho_p \cdot \tnode{$\cdot$}(z, x) + (1 - \rho_p) \cdot z \right)^{\infty x} \right) \\
        &\circ_z \tnode{\footnotesize pow} \left( \tnode{$\div$} \left( u, v \right), \tnode{$\gamma$} \right)
        \circ_u \tnode{$\cdot$} \left( \tnode{$q$}, \tnode{\footnotesize pow} \left( \tnode{$c$}, \tnode{$\alpha$} \right) \right)
        \circ_v \tnode{$+$} \left( \tnode{$1$}, \tnode{$\cdot$} \left( \tnode{$p$}, \tnode{\footnotesize pow} \left( \tnode{$c$}, \tnode{$\beta$} \right) \right) \right),
    \end{aligned}
\end{equation}
with $s_T, f_i, \gamma_{ij}, q_{ij}, \alpha_{ij}, p_{ij}, \beta_{ij} \in \Sigma^{(0)}$ marking continuous parameters $\theta_c$ to be inferred from data.
Note, that the probabilities $\rho_s$ and $\rho_p$ determine the size of the contained sums and products.

\begin{table*}[t]
    \caption{
    Regression of strain-energy densities for synthetic hyper-elastic materials. Best performance (avg.\ RMSE) bolded per dataset.}
    \label{tab:hyperelastic_results}
    \vskip 0.15in
    \begin{center}
    \begin{small}
    \begin{sc}
        \input{tables/hyperelastic/hyperelastic_minimal.tex}
    \end{sc}
    \end{small}
    \end{center}
    \vskip -0.1in 
\end{table*}

\paragraph{Datasets.}
\label{sec:application:synthetic_data}
We test generalization given a training dataset $\mathcal{D}_\text{train}$ of $20$ noisy observations on $10$ different isotherms with datasets
\begin{enumerate*}[label={(\roman*)}]
    \item from the same range as the training data $\mathcal{D}_\text{test1} \subseteq [20, 100]$,
    \item \textit{lower concentrations} $\mathcal{D}_\text{test2} \subseteq [0, 20]$, and
    \item \textit{higher concentrations} $\mathcal{D}_\text{test3} \subseteq [100, 150]$.
\end{enumerate*}

\paragraph{Results}
\label{sec:experimental_results}
in \Cref{tab:syn_isotherm_results} show the \gls{RMSE} statistics (mean $\pm$ std) taken over $50$ repetitions of the experiment for each combination of algorithm and dataset.
Here prior knowledge helps to generalize better in the majority ($17$ out of $30$) of the test set scenarios.
Especially in extrapolation, i.e.\ $\mathcal{D}_\text{test2}$ (9 out of 10), and $\mathcal{D}_\text{test3}$ (5 out of 10), our method performs better.

To better understand these results, we analyzed the complexity in terms of the shortest description lengths \citep{udrescu2020ai2} of the expressions found (see \cref{sec:complexity} for full details):
Even though the encoded domain knowledge is quite general (compare \Cref{fig:visual_results}\subref{fig:prte_prior}) it helps \gls{pRTE} to meet the right level of complexity of expressions (i.e.\ relative complexity $\approx 1$).
In contrast, \textit{SInDy} and \textit{GVAE} tend to generate too simple expressions and \textit{Genetic Programming} generates overly complex expressions 
causing \textit{unphysical} behavior (compare \Cref{fig:visual_results}\subref{fig:baseline_results}).
\Gls*{pRTE} yields an informed posterior (\Cref{fig:visual_results}\subref{fig:prte_posterior}) with estimates for uncertainties.


\subsection{Hyper-Elastic Materials}
Materials like rubber exhibit a non-linear behavior when stresses and strains are applied. 
A common model class for \textit{hyper-elastic materials}  are \textit{Ogden Models} \cite{ogden1972large}, that relate the \textit{strain energy density}
\begin{equation}
    W(\lambda_1, \lambda_2, \lambda_3) = \sum_{p=1}^N \frac{\mu_p}{\alpha_p} \left( \lambda_1^{\alpha_p} + \lambda_2^{\alpha_p} + \lambda_3^{\alpha_p} - 3 \right)
\end{equation}
to the eigenvalues of a deformation gradient tensor $C$, the \textit{principal stretches} $\lambda_1, \lambda_2, \lambda_3$ with parameters $\alpha_p, \mu_p$. 
We express this knowledge as an \gls{pRTE} $E_\text{hyp}$ with shared parameters \tnode{$\alpha$}, as
\begin{equation}
    \label{eq:hyperelastic_prte}
    \begin{aligned}
        &\left(\rho_s \cdot \tnode{$+$}(y, x) + (1 - \rho_s) \cdot y \right)^{\infty x}
        \circ_{y} \tnode{$\cdot$} \left( \tnode{$\div$}(\tnode{$\mu$}, \tnode{$\alpha$}), \tnode{-}\left( z, \tnode{3} \right) \right)
        \circ_{z} \tnode{$+$} \left(
        \begin{aligned}
            &\tnode{\footnotesize pow} ( \tnode{$\lambda_1$}, \tnode{$\alpha$} ), \\
            &\tnode{\footnotesize pow} ( \tnode{$\lambda_2$}, \tnode{$\alpha$} ), \\
            &\tnode{\footnotesize pow} ( \tnode{$\lambda_3$}, \tnode{$\alpha$} )
        \end{aligned}
        \right)
    \end{aligned}
\end{equation}
with $\rho_s$ denoting the probability of continuation.

\paragraph{Datasets.}
\label{sec:application:synthetic_data}
We test generalization given a dataset $\lambda_1, \lambda_2, \lambda_3 \in \mathcal{D}_\text{train} \subseteq [2.0, 3.0]$ of $20$ observations with test sets:
\begin{enumerate*}[label={(\roman*)}]
    \item $\mathcal{D}_\text{test1} \subseteq [2.0, 3.0]$ with i.i.d.\ samples from the same range as the training data,
    \item $\mathcal{D}_\text{test2} \subseteq [0.75, 1.75]$ with smaller stretches, and
    \item $\mathcal{D}_\text{test3} \subseteq [3.25, 4.25]$ with bigger stretches.
\end{enumerate*}

\paragraph{Results}
\Cref{tab:hyperelastic_results}
summarizes the generalization on three synthetic Ogden materials.
Even though the expressed knowledge is weak (i.e.\ little constraints compared to arbitrary polynomials) it helped to better generalize in all of our experiments. 
More complex material models which also can be expressed through \gls{pRTE} are discussed in the appendix. The advantage of our method is that it can be adapted to this kind of knowledge as well, while standard symbolic regression approaches can not.

\subsection{Limitations}
We see two major limitations of our work:
\begin{enumerate*}[label={(\roman*)}]
    \item If \textit{knowledge is too broad}, too many equations are possible and the structure selection becomes a random walk with bad mixing. Bayesian reasoning with uniform-like, uninformative priors performs rather poorly in benchmarks \cite{la2021contemporary}.
    \item If \textit{knowledge can not be expressed} with regular expression; e.g.\ tree languages with equivalences of subtrees \citep{rozenberg2012handbook} or function space properties like the derivative in a certain point, our method can not be applied.
\end{enumerate*}

%% file: tables/sorption/isotherm_minimal.tex
\begin{tabular}{ccrrrrr}
    \toprule
     &  & \multicolumn{5}{c}{\thead{RMSE (mean $\pm$ std) $[\frac{mg}{Kg}]$}} \\
    isotherm & data & pRTE & SInDy\citep{brunton2016discovering} & GVAE\citep{kusner2017grammar} & Genetic\citep{koza1990genetic} & DGSR\citep{holt2023deep} \\
    \midrule
    \multirow[c]{3}{*}{\shortstack[c]{Brunauer\\ Emmett\\ Teller}} & test1 & $1.96\;{\scriptstyle \pm0.42}$ & $\mathbf{0.26}$ & $1.64\;{\scriptstyle \pm0.51}$ & $5.42\;{\scriptstyle \pm7.39}$ & $3.22\;{\scriptstyle \pm0.04}$ \\
     & test2 & $\mathbf{3.09\;{\scriptstyle \pm1.59}}$ & $22.71$ & $3.41\;{\scriptstyle \pm1.06}$ & $110.02\;{\scriptstyle \pm77.22}$ & $184.83\;{\scriptstyle \pm1.23}$ \\
     & test3 & $30.81\;{\scriptstyle \pm3.45}$ & $100.45$ & $\mathbf{28.65\;{\scriptstyle \pm1.66}}$ & $116.04\;{\scriptstyle \pm38.12}$ & $127.32\;{\scriptstyle \pm32.11}$ \\
    \cline{1-7}
    \multirow[c]{3}{*}{\shortstack[c]{Freundlich}} & test1 & $0.41\;{\scriptstyle \pm0.06}$ & $\mathbf{0.30}$ & $0.55\;{\scriptstyle \pm0.61}$ & $1.29\;{\scriptstyle \pm1.31}$ & $6.05\;{\scriptstyle \pm0.00}$ \\
     & test2 & $\mathbf{1.11\;{\scriptstyle \pm0.94}}$ & $23.04$ & $3.09\;{\scriptstyle \pm2.10}$ & $95.12\;{\scriptstyle \pm87.55}$ & $59.86\;{\scriptstyle \pm4.03}$ \\
     & test3 & $\mathbf{0.41\;{\scriptstyle \pm0.06}}$ & $97.07$ & $0.99\;{\scriptstyle \pm1.00}$ & $23.48\;{\scriptstyle \pm30.79}$ & $27.79\;{\scriptstyle \pm0.00}$ \\
    \cline{1-7}
    \multirow[c]{3}{*}{\shortstack[c]{General\\ Freundlich}} & test1 & $0.42\;{\scriptstyle \pm0.04}$ & $\mathbf{0.13}$ & $0.25\;{\scriptstyle \pm0.15}$ & $2.32\;{\scriptstyle \pm7.22}$ & $0.47\;{\scriptstyle \pm0.00}$ \\
     & test2 & $\mathbf{3.49\;{\scriptstyle \pm1.74}}$ & $26.64$ & $5.03\;{\scriptstyle \pm0.50}$ & $74.66\;{\scriptstyle \pm55.46}$ & $36.88\;{\scriptstyle \pm0.00}$ \\
     & test3 & $0.47\;{\scriptstyle \pm0.04}$ & $49.09$ & $0.46\;{\scriptstyle \pm0.11}$ & $15.89\;{\scriptstyle \pm39.88}$ & $\mathbf{0.37\;{\scriptstyle \pm0.00}}$ \\
    \cline{1-7}
    \multirow[c]{3}{*}{\shortstack[c]{General\\ Langmuir\\ Freundlich}} & test1 & $0.38\;{\scriptstyle \pm0.04}$ & $\mathbf{0.14}$ & $0.27\;{\scriptstyle \pm0.35}$ & $1.08\;{\scriptstyle \pm1.45}$ & $38.00\;{\scriptstyle \pm111.04}$ \\
     & test2 & $\mathbf{1.05\;{\scriptstyle \pm0.85}}$ & $14.09$ & $2.68\;{\scriptstyle \pm1.64}$ & $72.81\;{\scriptstyle \pm60.38}$ & $68.60\;{\scriptstyle \pm110.25}$ \\
     & test3 & $\mathbf{0.39\;{\scriptstyle \pm0.04}}$ & $58.29$ & $0.58\;{\scriptstyle \pm0.55}$ & $12.09\;{\scriptstyle \pm24.29}$ & $38.26\;{\scriptstyle \pm108.41}$ \\
    \cline{1-7}
    \multirow[c]{3}{*}{\shortstack[c]{Langmuir}} & test1 & $0.64\;{\scriptstyle \pm0.03}$ & $15.62$ & $\mathbf{0.57\;{\scriptstyle \pm0.51}}$ & $2.16\;{\scriptstyle \pm3.88}$ & $2.26\;{\scriptstyle \pm0.41}$ \\
     & test2 & $\mathbf{5.08\;{\scriptstyle \pm3.64}}$ & $61.29$ & $24.95\;{\scriptstyle \pm46.15}$ & $98.92\;{\scriptstyle \pm78.10}$ & $11.84\;{\scriptstyle \pm6.36}$ \\
     & test3 & $\mathbf{0.96\;{\scriptstyle \pm0.18}}$ & $146.91$ & $1.54\;{\scriptstyle \pm0.73}$ & $27.33\;{\scriptstyle \pm44.64}$ & $1.18\;{\scriptstyle \pm0.71}$ \\
    \cline{1-7}
    \multirow[c]{3}{*}{\shortstack[c]{Modified\\ Langmuir}} & test1 & $\mathbf{0.86\;{\scriptstyle \pm0.07}}$ & $8.00$ & $1.54\;{\scriptstyle \pm0.76}$ & $2.41\;{\scriptstyle \pm4.07}$ & $2.23\;{\scriptstyle \pm0.82}$ \\
     & test2 & $\mathbf{4.02\;{\scriptstyle \pm2.83}}$ & $49.63$ & $9.30\;{\scriptstyle \pm4.04}$ & $111.70\;{\scriptstyle \pm67.50}$ & $149.61\;{\scriptstyle \pm33.69}$ \\
     & test3 & $1.44\;{\scriptstyle \pm0.30}$ & $74.26$ & $3.81\;{\scriptstyle \pm1.81}$ & $20.75\;{\scriptstyle \pm29.20}$ & $\mathbf{1.12\;{\scriptstyle \pm0.38}}$ \\
    \cline{1-7}
    \multirow[c]{3}{*}{\shortstack[c]{Redlich\\ Peterson}} & test1 & $0.56\;{\scriptstyle \pm0.03}$ & $1.60$ & $\mathbf{0.26\;{\scriptstyle \pm0.02}}$ & $1.94\;{\scriptstyle \pm5.29}$ & $1.00\;{\scriptstyle \pm0.20}$ \\
     & test2 & $\mathbf{5.23\;{\scriptstyle \pm1.95}}$ & $46.83$ & $8.62\;{\scriptstyle \pm0.06}$ & $77.35\;{\scriptstyle \pm55.06}$ & $46.26\;{\scriptstyle \pm0.03}$ \\
     & test3 & $0.73\;{\scriptstyle \pm0.13}$ & $17.75$ & $\mathbf{0.67\;{\scriptstyle \pm0.12}}$ & $25.97\;{\scriptstyle \pm46.69}$ & $4.30\;{\scriptstyle \pm0.25}$ \\
    \cline{1-7}
    \multirow[c]{3}{*}{\shortstack[c]{Toth}} & test1 & $\mathbf{0.62\;{\scriptstyle \pm0.04}}$ & $10.99$ & $0.64\;{\scriptstyle \pm0.50}$ & $2.61\;{\scriptstyle \pm7.40}$ & $1.56\;{\scriptstyle \pm0.02}$ \\
     & test2 & $\mathbf{4.29\;{\scriptstyle \pm1.51}}$ & $61.72$ & $9.17\;{\scriptstyle \pm2.57}$ & $99.00\;{\scriptstyle \pm79.29}$ & $23.64\;{\scriptstyle \pm44.23}$ \\
     & test3 & $\mathbf{0.85\;{\scriptstyle \pm0.16}}$ & $145.33$ & $1.45\;{\scriptstyle \pm0.55}$ & $29.37\;{\scriptstyle \pm50.25}$ & $6.46\;{\scriptstyle \pm0.02}$ \\
    \cline{1-7}
    \multirow[c]{3}{*}{\shortstack[c]{Two Site\\ Langmuir}} & test1 & $\mathbf{0.62\;{\scriptstyle \pm0.04}}$ & $0.65$ & $0.74\;{\scriptstyle \pm0.66}$ & $2.03\;{\scriptstyle \pm2.61}$ & $1.83\;{\scriptstyle \pm0.37}$ \\
     & test2 & $\mathbf{2.23\;{\scriptstyle \pm1.13}}$ & $48.51$ & $26.49\;{\scriptstyle \pm59.84}$ & $107.92\;{\scriptstyle \pm83.65}$ & $12.48\;{\scriptstyle \pm7.30}$ \\
     & test3 & $\mathbf{0.91\;{\scriptstyle \pm0.17}}$ & $115.14$ & $1.80\;{\scriptstyle \pm0.89}$ & $34.41\;{\scriptstyle \pm44.88}$ & $6.92\;{\scriptstyle \pm1.40}$ \\
    \cline{1-7}
    \multirow[c]{3}{*}{\shortstack[c]{Farley\\ Dzombak\\ Morel}} & test1 & $7.02 \;{\scriptstyle \pm 0.91}$ & $\mathbf{0.34}$ & $0.66\;{\scriptstyle \pm0.02}$ & $5.85\;{\scriptstyle \pm6.31}$ & $3.79\;{\scriptstyle \pm0.17}$ \\
     & test2 & $39.24 \;{\scriptstyle\pm 20.36}$ & $50.70$ & $10.04\;{\scriptstyle \pm0.65}$ & $96.78\;{\scriptstyle \pm58.85}$ & $\mathbf{8.41\;{\scriptstyle \pm0.31}}$ \\
     & test3 & $1446.22 \;{\scriptstyle\pm 0.55}$ & $1459.91$ & $\mathbf{7.51\;{\scriptstyle \pm0.01}}$ & $1449.11\;{\scriptstyle \pm4.14}$ & $1449.82\;{\scriptstyle \pm0.34}$ \\
    \bottomrule
    \end{tabular}

%% file: tables/hyperelastic/hyperelastic_minimal.tex
\begin{tabular}{lllllll}
    \toprule
     & & \multicolumn{5}{c}{\thead{RMSE (mean $\pm$ std) $[J]$}} \\
    material & data & pRTE & SInDy\citep{brunton2016discovering} & Genetic\citep{koza1990genetic} & DGSR\citep{holt2023deep} & GVAE\citep{kusner2017grammar} \\
    \midrule
    \multirow[t]{3}{*}{ogden-1} & test1 & $\mathbf{0.007\;{\scriptstyle  \pm0.008}}$ & $0.127\;{\scriptstyle  \pm0.007}$ & $0.131\;{\scriptstyle  \pm0.035}$ & $0.131\;{\scriptstyle  \pm0.000}$ & $0.080\;{\scriptstyle  \pm0.020}$ \\
     & test2 & $\mathbf{0.038\;{\scriptstyle  \pm0.034}}$ & $0.531\;{\scriptstyle  \pm0.016}$ & $0.366\;{\scriptstyle  \pm0.229}$ & $0.629\;{\scriptstyle  \pm0.069}$ & $0.571\;{\scriptstyle  \pm0.328}$ \\
     & test3 & $\mathbf{0.237\;{\scriptstyle  \pm0.226}}$ & $1.774\;{\scriptstyle  \pm0.141}$ & $0.408\;{\scriptstyle  \pm0.187}$ & $0.299\;{\scriptstyle  \pm0.000}$ & $0.619\;{\scriptstyle  \pm0.327}$ \\
    \cline{1-7}
    \multirow[t]{3}{*}{ogden-2} & test1 & $\mathbf{0.005\;{\scriptstyle  \pm0.004}}$ & $0.506\;{\scriptstyle  \pm0.023}$ & $0.486\;{\scriptstyle  \pm0.058}$ & $0.551\;{\scriptstyle  \pm0.006}$ & $0.287\;{\scriptstyle  \pm0.037}$ \\
     & test2 & $\mathbf{0.036\;{\scriptstyle  \pm0.026}}$ & $1.093\;{\scriptstyle  \pm0.081}$ & $6.249\;{\scriptstyle  \pm15.448}$ & $0.861\;{\scriptstyle  \pm0.172}$ & $0.974\;{\scriptstyle  \pm0.351}$ \\
     & test3 & $\mathbf{0.280\;{\scriptstyle  \pm0.181}}$ & $5.321\;{\scriptstyle  \pm0.810}$ & $1.672\;{\scriptstyle  \pm0.780}$ & $1.879\;{\scriptstyle  \pm0.248}$ & $2.102\;{\scriptstyle  \pm0.625}$ \\
    \cline{1-7}
    \multirow[t]{3}{*}{ogden-3} & test1 & $\mathbf{0.010\;{\scriptstyle  \pm0.013}}$ & $0.079\;{\scriptstyle  \pm0.003}$ & $0.071\;{\scriptstyle  \pm0.020}$ & $0.084\;{\scriptstyle  \pm0.001}$ & $0.056\;{\scriptstyle  \pm0.004}$ \\
     & test2 & $\mathbf{0.024\;{\scriptstyle  \pm0.026}}$ & $0.300\;{\scriptstyle  \pm0.010}$ & $0.242\;{\scriptstyle  \pm0.136}$ & $0.495\;{\scriptstyle  \pm0.084}$ & $0.372\;{\scriptstyle  \pm0.357}$ \\
     & test3 & $\mathbf{0.151\;{\scriptstyle  \pm0.121}}$ & $1.030\;{\scriptstyle  \pm0.083}$ & $0.331\;{\scriptstyle  \pm0.101}$ & $0.202\;{\scriptstyle  \pm0.009}$ & $0.326\;{\scriptstyle  \pm0.128}$ \\
    \cline{1-7}
    \bottomrule
\end{tabular}

%% file: sections/conclusion.tex
\section{Conclusion}
\label{sec:conclusion}
We showed that the use of tree languages comes with several advantages to encoding scientific prior knowledge as a prior distribution for symbolic regressions:
\begin{enumerate*}[label={(\roman*)}]
    \item compact notation with regular tree expressions,
    \item closedness under boolean operations
\end{enumerate*}.
Our scientific case study showed that reasoning with existing but general prior knowledge helped to better generalize sorption processes in soil science $17$ out of $30$ test set scenarios. When applied to the identification of hyper-elastic materials our method consistently outperformed all considered baselines.

%% file: formal/acknowledgement.tex
\section*{Acknowledgements}
The research presented in this paper has been
funded by Deutsche Forschungsgemeinschaft (DFG, German Research Foundation) under Germany's Excellence Strategy - EXC 2075 – 390740016 and
the Ministry of Science, Research and
the Art Baden-Wuerttemberg via the Artificial Intelligence Software Academy
(AISA).
We acknowledge the support by the Stuttgart Center for Simulation Science (SimTech).

%% file: sections/appendix/implementation/logdetjacobian.tex
The constructions of the differentiable maps $f$ to model jumps between parameter spaces of different sizes are taken from \citet{jin2019bayesian}.
However, the actual computation of the determinants of the jacobians that is needed for an implementation of their method is skipped by the authors.
Thus we provide this missing derivation for general jumps, considering both \textit{expansion} (see \cref{sec:det_expansion}) and \textit{shrinkage} (see \cref{sec:det_shrinkage}) of parameter vectors $\theta$.

\subsubsection{Expansion of Parameters}
\label{sec:det_expansion}
Consider $f: (\theta, u) \mapsto (\theta^*, u^*)$ with $u = \begin{pmatrix} u_\theta \\ u_n \end{pmatrix}$ mapping to $\theta^* = \begin{pmatrix} \frac{\theta + u_\theta}{2} \\ u_n \end{pmatrix}$ and $u^*=\frac{\theta - u_\theta}{2}$ that increases the number of parameters in a symbolic expression.
If the parameter vector expands from size $n$ to size $n^*$ one can determine the determinant of the Jacobian as follows:

\begin{align}
    \det \Big(\frac{\partial (\theta^*, u^*)}{\partial (\theta, u)} \Big) &= \det
    \begin{pmatrix}
        \frac{\partial \theta^*}{\partial \theta } & \frac{\partial \theta^*}{\partial u} \\ 
        \frac{\partial u^*}{\partial \theta} & \frac{\partial u^*}{\partial u}
    \end{pmatrix} \\
    &= {\frac{1}{2}}^{(n + n^*)} \cdot \det
    \left(
    \begin{array}{c | c}
        \begin{matrix}
            1 & & \\
            & \ddots & \\
            & & 1 \\
            0 & \cdots & 0 \\
            \vdots & \ddots & \vdots \\
            0 & \cdots & 0
        \end{matrix} &
        \begin{matrix}
            1 & & & &\\
            & \ddots & & &\\
            & & 1 & & \\
            & & & 2 & \\
            & & & & \ddots \\
            & & & & & 2
        \end{matrix} \\
        \hline
        \begin{matrix}
            1 & & \\
            & \ddots & \\
            & & 1
        \end{matrix} &
        \begin{matrix}
            -1 & & & 0 & \cdots & 0 \\
            & \ddots & & \vdots & \ddots & \vdots \\
            & & -1 & 0 & \cdots & 0
        \end{matrix}
    \end{array}
    \right) \\
    &= {\frac{1}{2}}^{(n + n^*)} \cdot 2^{(n^* - n)} \det
    \left(
    \begin{array}{c | c}
        \begin{matrix}
            1 & & \\
            & \ddots & \\
            & & 1
        \end{matrix} &
        \begin{matrix}
            1 & & \\
            & \ddots & \\
            & & 1
        \end{matrix} \\
        \hline
        \begin{matrix}
            1 & & \\
            & \ddots & \\
            & & 1
        \end{matrix} &
        \begin{matrix}
            -1 & &  \\
            & \ddots &  \\
            & & -1
        \end{matrix}
    \end{array}
    \right) \\
    &= {\frac{1}{2}}^{(n + n^*)} \cdot 2^{(n^* - n)} \det\Big(I_n \Big) \cdot \det \Big(I_n - I_n I_n^{-1} \cdot (-1) \cdot I_n \Big) \\
    &= {\frac{1}{2}}^{(n + n^*)} \cdot 2^{(n^* - n)} \cdot 2^n = 2^{-n}
\end{align}

\subsubsection{Shrinkage of Parameters}
\label{sec:det_shrinkage}
Consider $f: (\theta, u) \mapsto (\theta^*, u^*)$ with $\theta = \begin{pmatrix} \theta^0 \\ \theta^d \end{pmatrix}$ mapping to $\theta^* = \theta^0 + u$ and $u^*= \begin{pmatrix}\theta^0 - u \\ \theta^d
\end{pmatrix}$ that reduces the number of parameters in a symbolic expression.
If the parameter vector shrinks from size $n$ to size $n^*$ one can determine the determinant of the Jacobian as follows:

\begin{align}
    \det \Big(\frac{\partial (\theta^*, u^*)}{\partial (\theta, u)} \Big) &= \det
    \begin{pmatrix}
        \frac{\partial \theta^*}{\partial \theta } & \frac{\partial \theta^*}{\partial u} \\ 
        \frac{\partial u^*}{\partial \theta} & \frac{\partial u^*}{\partial u}
    \end{pmatrix} \\
    &= \det
    \left(
    \begin{array}{c | c}
        \begin{matrix}
            1 & & & 0 & \cdots & 0 \\
            & \ddots & & \vdots & \ddots & \vdots \\
            & & 1 & 0 & \cdots & 0
        \end{matrix} &
        \begin{matrix}
            1 & & \\
            & \ddots & \\
            & & 1
        \end{matrix} \\
        \hline
        \begin{matrix}
            1 & & \\
            & \\
            & & \ddots & & \\
            & \\
            & & & & 1
        \end{matrix} &
        \begin{matrix}
            -1 & & \\
            & \ddots & \\
            & & -1 \\
            0 & \cdots & 0 \\
            \vdots & \ddots & \vdots \\
            0 & \cdots & 0
        \end{matrix}
    \end{array}
    \right) \\
    &= \det
    \left(
    \begin{array}{c | c}
        \begin{matrix}
            1 & & \\
            & \ddots & \\
            & & 1
        \end{matrix} &
        \begin{matrix}
            1 & & \\
            & \ddots & \\
            & & 1
        \end{matrix} \\
        \hline
        \begin{matrix}
            1 & & \\
            & \ddots & \\
            & & 1
        \end{matrix} &
        \begin{matrix}
            -1 & &  \\
            & \ddots &  \\
            & & -1
        \end{matrix}
    \end{array}
    \right) \\
    &= \det\Big(I_{n^*} \Big) \cdot \det \Big(I_{n^*} - I_{n^*} I_{n^*}^{-1} \cdot (-1) \cdot I_{n^*} \Big) \\
    &=  2^{n^*}
\end{align}

%% file: sections/appendix/implementation/hyperparameters.tex
The posterior distributions of symbolic solutions are approximated through $5000$ samples taken after $15000$ burn-in steps. 
The samples are thinned by a factor of $100$ to $50$ effective samples. 
Furthermore, we consistently chose $\lambda = 1$, and $\rho_s = \rho_p = 0.1$ for all experiments.

The priors for the continuous parameters in the pRTEs are chosen $\mathcal{N}(1, 0.1)$. 
Here, the expectation of one avoids meaningless exponents and divisions by zero in too many proposals.
Only the prior for $S_T$ is chosen $\text{Exp}(0.15)$ to scale the expressions a-priori.
With these choices, the prior is not informative. Compare the function scatters in \cref{fig:prte_prior}.

%% file: sections/appendix/implementation/context.tex
Incremental steps to an existing equation allow us to make better proposals and achieve better local connectivity of the Markov chain.
To that end, the partial evaluation of densities allows us to propose \emph{local changes} to a tree structure $t$ at hand, e.g.\ by only replacing a subtree of an already promising equation. 
To achieve this, one needs to understand the evaluation of a tree series as the full marginal inference in the factor graphs. Not marginalizing a certain state variable $q_r$ yields a distribution over states $\mu_r \in \Delta(Q)$ for the tree position $r \in D $. This distribution describes the possible states for a \gls{PTA}, given a partial tree $t \in $ as \textit{context}. Again, we use a special symbol $t(r) = \tnode{$?$} \in \Sigma_0$ to mark the position of the context state query, as shown in \Cref{fig:context_resampling}.
We relax the categorical distribution $\mu_r$ to a Boltzmann distribution $p_B(q_r = i) \propto \exp{\frac{\log \mu_r(i)}{\tau}} \; \forall i \in \{1, \ldots, N\}$ with a temperature $\tau$ that controls the exploration of a-priori unlikely states.
Sampling $q_r \sim p_B$ and regrowing the tree $t$ from position $r$ (compare \cref{sec:method:prior_sampling}) yields a new proposal tree structure $t'$. 
This is similar to the local steps in \cite{jin2019bayesian}) but proposals are consistent with the prior knowledge by excluding states with zero probability in $\mu_r$.

\begin{figure}
     \centering
     \begin{subfigure}[b]{0.4\linewidth}
         \centering
         \scalebox{0.7}{\input{figures/context_resampling/context_tree}}
         \caption{$t_{\leq r} \in T_{\Sigma \cup \{?\}}$}
         \label{fig:context_tree}
     \end{subfigure}
     \hfill
     \begin{subfigure}[b]{0.55\linewidth}
         \centering
         \scalebox{0.7}{\input{figures/context_resampling/factor_graph}}
         \caption{resulting factor graph.}
         \label{fig:context_factor_graph}
     \end{subfigure}
    \caption{Marginal inference for $p(q_4)$ in a factor graph (\subref{fig:context_factor_graph}) to re-sample the context $t_{\leq r}$ of a tree $t$ in position $r \in D$ marked by $? \in \Sigma^{(0)}$. With the inferred state $q_4$ a new sub-tree can be grown.}
    \label{fig:context_resampling}
\end{figure}

%% file: figures/context_resampling/context_tree.tex
\definecolor{blue}{rgb}{0.0, 0.45, 0.73}
\begin{tikzpicture}

\tikzstyle symbolic node=[draw, blue, thick, fill=blue!10!white, circle, minimum size=2em, text=black];
\tikzstyle symbolic link=[draw, blue, very thick];

\node[symbolic node] (plus1) {$+$};
\node[symbolic node, below= 0.5em of plus1, xshift= -1.5em] (plus2) {$+$};
\node[symbolic node, below= 0.5em of plus1, xshift= 2em] (twelve) {$c$};

\node[symbolic node, below= 0.5em of plus2, xshift= -1.5em] (sq1) {$\cdot^{2}$};
\node[symbolic node, below= 0.5em of plus2, xshift= 1.5em] (sq2) {$?$};

\node[symbolic node, below= 0.5em of sq1] (minus) {$-$};
\node[symbolic node, below= 0.5em of minus, xshift=-1.5em] (x) {$x$};
\node[symbolic node, below= 0.5em of minus, xshift=1.5em] (seven) {$c$};

\draw[symbolic link] (plus1) -- (plus2);
\draw[symbolic link] (plus1) -- (twelve);
\draw[symbolic link] (plus2) -- (sq1);
\draw[symbolic link] (plus2) -- (sq2);
\draw[symbolic link] (sq1) -- (minus);
\draw[symbolic link] (x) -- (minus);
\draw[symbolic link] (seven) -- (minus);

\end{tikzpicture}

%% file: figures/context_resampling/factor_graph.tex
\begin{tikzpicture}
    \node[latent] (q0) {$q_0$};
    \node[latent, below= 0.5em of q0, xshift= -1.5em] (q1) {$q_1$};
    \node[latent, below= 0.5em of q0, xshift= 1.5em] (q2) {$q_2$};
    \node[latent, below= 0.5em of q1, xshift= -2em] (q3) {$q_3$};
    \node[latent, below= 0.5em of q1, xshift= 5em] (q4) {$q_4$};
    \node[latent, below= 0.5em of q3] (q5) {$q_5$};
    \node[latent, below= 0.5em of q5, xshift= -1.5em] (q7) {$q_7$};
    \node[latent, below= 0.5em of q5, xshift= 1.5em] (q8) {$q_8$};

    \node[obs, right= 1em of q0] (x0) {$x_0$};
    \node[obs, left= 1em of q1] (x1) {$x_1$};
    \node[obs, right= 1em of q2] (x2) {$x_2$};
    \node[obs, left= 1em of q3] (x3) {$x_3$};
    \node[obs, left= 1em of q5] (x5) {$x_5$};
    \node[obs, left= 1em of q7] (x7) {$x_7$};
    \node[obs, right= 0.25em of q8, yshift=2em] (x8) {$x_8$};
    
    \factor[-, left= 0.5em of q0] {} {left:$\phi_0$} {q0} {};
    \factor[-, below= 0.25em of q0] {} {} {x0, q0, q1, q2} {};
    \factor[-, below= 0.5em of q1] {} {below:$\phi_1$} {x1, q1, q3, q4} {};
    \factor[-, right= 0.25em of q2] {} {} {x2, q2} {};
    \factor[-, below= 0.0em of q3, xshift=-1em] {} {} {x3, q3, q5} {};
    \factor[-, below= 0.25em of q5] {} {} {x5, q5, q7, q8} {};
    \factor[-, left= 0.25em of q7] {} {} {x7, q7} {};
    \factor[-, right= 0.25em of q8] {} {} {x8, q8} {};

\end{tikzpicture}

%% file: sections/appendix/implementation/baselines.tex
For all of our experiments, we used the following baseline implementations.

\subsubsection{Genetic Programming (GP)}
Here, we use the implementation of \textit{genetic.SymbolicRegressor} provided by \textit{gplearn} package with $250$ generations for the genetic evolution. All other hyper-parameters are left at the default values.

\subsubsection{Sparse Identification (SInDy)}
Here, we solve an ordinary \textit{LASSO} regression problem (using the implementation in \textit{Sci-kit Learn} with $10^7$ iterations) by interpreting the feature matrix $\Phi(\mathbf{X})$ as a \textit{library of terms} and enforced sparsity on the solution
\begin{align}
    \hat{\xi} = \arg \min_{\xi} \;\; (\mathbf{Y} - \Phi(\mathbf{X})\xi)^T (\mathbf{Y} - \Phi(\mathbf{X})\xi) + \alpha \| \xi \|_1
\end{align}
 with a hyper-parameter $\alpha= 0.1$. This allows for a sparse identification of the actual symbolic expression in a postprocessing step. Let $\mathbf{X} \in \mathbb{R}^{n \times m}$ be $m$ data samples with dimension $n$ each. Then $\mathbf{X}^{P_i}$ denotes the matrix with all polynomial terms of degree $i$ for each of the $m$ samples. The library 
\begin{align}
    \Phi(\mathbf{X}) = 
    \begin{pmatrix} 
    | & | & | & | & | & | & | & | & | \\
    1 & \mathbf{X} & \mathbf{X}^{P_i} &  \sin(\mathbf{X}) & \cos(\mathbf{X}) &  \tan(\mathbf{X}) &  \tanh(\mathbf{X}) & \frac{1}{X} & \sqrt{(X)} & \ldots \\
    | & | & | & | & | & | & | & | & |
    \end{pmatrix}
\end{align}
is composed from stacked matrices $\mathbf{X}$ in combination with functions of choice. We include polynomials up to a degree of $i \leq 6$ and also $\log(X + \epsilon_1)$ $\exp(X / \epsilon_1 + \epsilon_2)$ with paramaters $\epsilon$ to avoid exploding or ill-defined values.

\subsubsection{Grammar Variational Autoencoder (GVAE)}
We implemented a variant of the autoencoder that can deal with continuous variables inside of a symbolic expression. For the training, we put a special symbol in the grammar which marks the position of parameters and trained on $10^5$ random equations with a batch size of $2048$ until convergence after roughly $20$ epochs. At inference time, to compute the acquisition function of the latent GP, we first decode a latent code $z \in \mathbb{R}^{16}$ to a symbolic structure and use $100$ steps of a \textit{BFGS} optimizer with learning rate $10^{-2}$ to solve for the continuous parameters first. All other parameters are left at the default values from \cite{kusner2017grammar}.

\subsubsection{Deep Generative Symbolic Regression (DGSR)}
Here, we used the implementation released by \citet{holt2023deep} (\url{https://github.com/samholt/DeepGenerativeSymbolicRegression}) and used the provided pre-trained checkpoints of the transformer model for our inference. For the sorption experiments, the checkpoint \textit{"1-covar-koza"} for a single independent variable is used. For the hyper-elastic material experiments we respectively use the checkpoint \textit{"3-covar-koza"}. All other hyperparameters and configs are left at their default values.

%% file: sections/appendix/dataset.tex
\subsection{The Synthetic Sorption Isotherm Dataset}
\label{sec:synthetic_dataset}
As described in \cref{sec:application:synthetic_data}, we chose standard isotherm equations from literature to generate challenging symbolic regression tasks for the scientific domain of soil science. The ground truth equation is only used to generate the data and is presented to the regression algorithms at no time. After choosing an equation from the literature, we sample values for the parameters that occur inside it from the following priors:

\begin{align*}
    s_T &\sim \operatorname{Exp}(\lambda = 0.015) \\
    k, k_1, k_3, f_1, f_2, \alpha &\sim \operatorname{Exp}(\lambda = 4) \\
    k_2 &\sim \operatorname{Exp}(\lambda = 100) \\
    X, X_c &\sim \operatorname{Exp}(\lambda = 0.03) \\
    K_F &\sim \operatorname{Exp}(\lambda = 0.05) \\
\end{align*}
For \textit{Redlich Peterson} and \textit{Toth} we chose:
\begin{align*}
    \alpha &\sim \operatorname{Exp}(\lambda = 0.75) \\
\end{align*}
and for \textit{Brunauer Emmett Teller} we chose:
\begin{align*}
    k_1 &\sim \operatorname{Exp}(\lambda = 0.25) \\
    k_2 &\sim \operatorname{Exp}(\lambda = 4) \\
    k_3 &\sim \operatorname{Exp}(\lambda = 100) \\
\end{align*}
and for \textit{Two Site Langmuir} we chose:
\begin{align*}
    k_1, k_2 &\sim \operatorname{Exp}(\lambda = 8).
\end{align*}

Afterward, the inputs $c$ to the equations are uniformly sampled according to the intervals in $\mathcal{D}_\text{train}, \mathcal{D}_\text{test1}, \mathcal{D}_\text{test2}, \mathcal{D}_\text{test3}$ and the outputs $s$ are corrupted with white noise $\epsilon_i \sim \mathcal{N}(\mu =0, \sigma=0.1)$. These input/output pairs $(c_i, s_i + \epsilon_i)$ form the dataset for our symbolic regression task.
Furthermore, we fix the random seeds of the dataset generation (with an individual seed for each dataset and parameter sampling) to make the setup reproducible over repetitions and comparison against baselines.
The implementation of the data generation process is included in the code accompanying this submission.

\subsection{Hyper-Elastic Material Models}

For the experiment, we choose a ground truth \textit{Ogden Hyper-Elastic Material Model} with $N=2$, $\alpha = (1.75, 2.5)^T$, and $\mu = (1.5, 0.1)^T$. The input principal stretches $\lambda_1, \lambda_2, \lambda_3$ to this equation are uniformly sampled according to the intervals in $\mathcal{D}_\text{train}, \mathcal{D}_\text{test1}, \mathcal{D}_\text{test2}, \mathcal{D}_\text{test3}$ and the outputs $w$ are corrupted with white noise $\epsilon_i \sim \mathcal{N}(\mu =0, \sigma=0.01)$. These input/output pairs $(\lambda_{1i}, \lambda_{2i}, \lambda_{3i}, w_i + \epsilon_i)$ form the dataset for our symbolic regression task.

%% file: sections/appendix/examples/rebuttal.tex
This section provides additional (and longer) examples for the definition of trees, symbolic expressions and probability distributions over it.
First, \cref{ex:extended_prte_example} details how a tree is sampled from an \gls{pRTE}.
\begin{example}[Extended pRTE example]
    \label{ex:extended_prte_example}
    \textbf{All trees with at least one \tnode{g}:} \newline
    In this example we consider an abstract \textit{ranked alphabet} $\Sigma = \Sigma^{(0)} \cup \Sigma^{(1)} \cup \Sigma^{(2)}$ with 
    \begin{align*}
        \Sigma^{(0)} &:= \left\{ \tnode{a}, \tnode{b} \right\} &
        \Sigma^{(1)} &:= \left\{ \tnode{g} \right\} &
        \Sigma^{(2)} &:= \left\{ \tnode{f} \right\}
    \end{align*}
    This alphabet implies a set of trees $T_\Sigma$ over a ranked alphabet $\Sigma$, with e.g.\
    \begin{align*}
        &\tnode{f}(\tnode{a}, \tnode{b}) \in T_\Sigma & 
        &\tnode{a} \in T_\Sigma &
        &\tnode{g}(\tnode{b}) \in T_\Sigma &
        &\tnode{f}(\tnode{a}, \tnode{g}(\tnode{b})) \in T_\Sigma
    \end{align*}
    In the next step, we want to express some experts' domain knowledge for this artificial application. Let's assume that we know (for some reason) that the equation we are looking for must contain at least one $\tnode{g}$ symbol. Let's craft a regular expression ($E_1$) that equivalently expresses this knowledge:
    \begin{equation*}
        \begin{aligned}
            E_1 = \left(\tnode{f}(x,y) + \tnode{f}(y, x) + \tnode{g}(x) \right)^{\infty y}
            \circ_x \left( \tnode{f}(x,x) + \tnode{g}(x) + \tnode{a} + \tnode{b} \right)^{\infty x}
        \end{aligned}
    \end{equation*}
    This expression can be used to generate the tree $\tnode{f}(\tnode{a}, \tnode{g}(\tnode{b})) \in T_\Sigma$, let's see step by step how:
    \begin{align*}
        y \rightarrow \tnode{f}(x, y) \rightarrow \tnode{f}(x, \tnode{g}(x)) \rightarrow \tnode{f}(\tnode{a}, \tnode{g}(x)) \rightarrow \tnode{f}(\tnode{a}, \tnode{g}(\tnode{b}))
    \end{align*}
    We start with a variable $y \in V$ which is not part of the alphabet $y \notin \Sigma$ but a special symbol. All intermediate steps of a generation are trees $t \in T_{\Sigma \cup V}$ and the generation ends when $t \in T_{\Sigma}$. First, $\infty y$ says that all $y$ must be iteratively replaced with the inner expression $\tnode{f}(x,y) + \tnode{f}(y, x) + \tnode{g}(x)$. From this choice $+$ we choose $\tnode{f}(x,y)$. Since it still contains a $y$ the iteration continues. Next, we choose the y to be replaced by $\tnode{g}(x)$; the iteration stops. Now, $\circ_x$  replaces all occurrences of $x$ by a sample from the subexpression to its right. The left $x$ gets replaces by $\tnode{a}$ and the right $x$ by $\tnode{b}$.
    More generally, these rules define a tree language $L \subseteq T_\Sigma$, with e.g.\
    \begin{align*}
        &\tnode{f}(\tnode{a}, \tnode{b}) \notin L & 
        &\tnode{a} \notin L &
        &\tnode{g}(\tnode{b}) \in L &
        &\tnode{f}(\tnode{a}, \tnode{g}(\tnode{b})) \in L
    \end{align*}
    Let's turn it into a probabilistic regular tree expression ($E_2$) by making the choice operator $+$ probabilistic. The weights in front of each element determine the probability of by chosen:
    \begin{equation*}
        \begin{aligned}
            E_2 = \left(\frac{1}{3}\tnode{f}(x,y) + \frac{1}{3} \tnode{f}(y, x) + \frac{1}{3} \tnode{g}(x) \right)^{\infty y}
            \circ_x \left( \frac{1}{4} \tnode{f}(x,x) + \frac{1}{4} \tnode{g}(x) + \frac{1}{4}\tnode{a} + \frac{1}{4} \tnode{b} \right)^{\infty x}
        \end{aligned}
    \end{equation*}
    and thus with probabilities $\frac{1}{3} \frac{1}{3} \frac{1}{4} \frac{1}{4} = \frac{1}{144}$ during sampling, i.e.\
    \begin{align*}
        y \overset{\frac{1}{3}}{\longrightarrow} \tnode{f}(x, y) 
        \overset{\frac{1}{3}}{\longrightarrow} \tnode{f}(x, \tnode{g}(x)) 
        \overset{\frac{1}{4}}{\longrightarrow} \tnode{f}(\tnode{a}, \tnode{g}(x)) \overset{\frac{1}{4}}{\longrightarrow} \tnode{f}(\tnode{a}, \tnode{g}(\tnode{b}))
    \end{align*}
    This expression thus implies a probability distribution $p: T_\Sigma \rightarrow [0,1]$, with e.g.\ 
    \begin{align*}
        p\left(\tnode{f}(\tnode{a}, \tnode{b})\right) &= 0 & 
        p\left(\tnode{a}\right) &= 0 &
        p\left(\tnode{g}(\tnode{b})\right) &= \frac{1}{12} &
        p\left(\tnode{f}(\tnode{a}, \tnode{g}(\tnode{b}))\right) &= \frac{1}{144}
    \end{align*}
    \end{example}
    
    The purpose of the following \cref{ex:symbolic_expression} is to show how a tree structure (e.g.\ sampled following the procedure from the previous example) is enriched with tuples of parameters to form a symbolic expression that can be evaluated.
    
    \begin{example}[Symbolic Expression Definition]
        \label{ex:symbolic_expression}
        In this example we consider the \textit{ranked alphabet} $\Sigma = \Sigma^{(0)} \cup \Sigma^{(1)} \cup \Sigma^{(2)}$ with special symbols $\tnode{c}, \tnode{d}$ and
    \begin{align*}
        \Sigma^{(0)} &:= \left\{ \tnode{x}, \tnode{c}, \tnode{d} \right\} &
        \Sigma^{(1)} &:= \left\{ \tnode{$\sin$} \right\} &
        \Sigma^{(2)} &:= \left\{ \tnode{+}, \tnode{$\cdot$}, \tnode{\footnotesize pow} \right\}.
    \end{align*}
    A valid symbolic expression $\mathfrak{e} = (t, \theta_d, \theta_c) \in \mathfrak{E}$ is defined by $t = \tnode{\footnotesize pow}(\tnode{+}(\tnode{$\cdot$}(\tnode{c},\tnode{$\sin$}(\tnode{x})), \tnode{c}), \tnode{d})$, $\operatorname{pos}_c = \{ 01, 0000 \} \subseteq D$, $\operatorname{pos}_d = \{ 1 \} \subseteq D$, $\Theta_c(t) := \mathbb{R}^{2}$, $\Theta_d(t) := \mathbb{Q}$ and e.g.\ $\theta_c = (0.25, 1.06)^T \in \Theta_c(t)$, $\theta_d = 2 \in \mathbb{Q}$. This yields the symbolic expression $(\sin(0.25 x) + 1.06)^2$.
    \end{example}

%% file: sections/appendix/examples/materials.tex
In literature exist several hyper-elastic material models which can be expressed as prior knowledge with our proposed method as \gls{pRTE} for Bayesian inference. These, non-exclusively compromise \textit{Neo Hookean Solids} (\cref{sec:neo_hookean}), \textit{Mooney Rivlin Solids} (\cref{sec:mooney_rivlin}), and \textit{Generalized Rivlin Solids} (\cref{sec:general_rivlin}).

\subsubsection{Neo Hookean Solid}
\label{sec:neo_hookean}
This is a particular case for the \textit{Ogden Material Model} as presented in the main paper (see \cref{eq:hyperelastic_prte}) with $N=1, \alpha_1=2, \mu_1= C_1 \alpha_1$ which yields
\begin{equation}
    W = C_1 \cdot (I_1 - 3) = C_1 \cdot (\lambda_1^2 + \lambda_2^2 + \lambda_3^2 - 3)
\end{equation}
expressed as
\begin{equation}
    \label{eq:neo_hookean_prte}
    \begin{aligned}
        E_\text{hook} = \tnode{$\cdot$} \left( \tnode{$C_1$},
        \tnode{-}\left(
        \tnode{$+$} \left( \tnode{\footnotesize pow} ( \tnode{$\lambda_1$}, \tnode{$2$} ),
        \tnode{\footnotesize pow} ( \tnode{$\lambda_2$}, \tnode{$2$} ),
        \tnode{\footnotesize pow} ( \tnode{$\lambda_3$}, \tnode{$2$} )\right)
        , \tnode{3} \right) \right)
    \end{aligned}
\end{equation}
with a (real) parameter for the occurrence of symbol \tnode{$C_1$}.

\subsubsection{Mooney Rivlin Solid}
\label{sec:mooney_rivlin}
Another common model from literature is the \textit{Mooney Rivlin Solid}
\begin{equation}
    W = C_1 \cdot (\bar{I}_1 - 3) + C_2 \cdot (\bar{I}_2 - 3)
\end{equation}
with
\begin{align*}
    \bar{I}_1 &:= J^{-\frac{2}{3}} \cdot I_1 & \bar{I}_2 &:= J^{-\frac{4}{3}} \cdot I_2 & J &= \lambda_1 \lambda_2 \lambda_3 \\
    I_1 &:= \lambda_1^2 + \lambda_2^2 + \lambda_3^2 & I_2 &:= \lambda_1^2 \lambda_2^2 + \lambda_2^2 \lambda_3^2 + \lambda_3^2 \lambda_1^2 & 
\end{align*}
expressed as
\begin{equation}
    \label{eq:mooney_rivlin_prte}
    \begin{aligned}
        E_{\bar{I}_1} &:= \tnode{$\cdot$} \left( \tnode{\footnotesize pow}\left( E_J, \tnode{$-\frac{2}{3}$}\right) , z \right) \circ_z  \tnode{$+$} \left( \tnode{\footnotesize pow} ( \tnode{$\lambda_1$}, \tnode{$2$} ),
        \tnode{\footnotesize pow} ( \tnode{$\lambda_2$}, \tnode{$2$} ),
        \tnode{\footnotesize pow} ( \tnode{$\lambda_3$}, \tnode{$2$} )\right)\\
        E_{\bar{I}_2} &:= \tnode{$\cdot$} \left( \tnode{\footnotesize pow}\left( E_J, \tnode{$-\frac{4}{3}$}\right) , \tnode{$+$} \left( z_1, z_2, z_3\right) \right) \\
        &\circ_{z_1} \tnode{$\cdot$} \left( \tnode{\footnotesize pow} ( \tnode{$\lambda_1$}, \tnode{$2$} ), \tnode{\footnotesize pow} ( \tnode{$\lambda_2$}, \tnode{$2$} ) \right)\\
        &\circ_{z_2} \tnode{$\cdot$} \left( \tnode{\footnotesize pow} ( \tnode{$\lambda_2$}, \tnode{$2$} ), \tnode{\footnotesize pow} ( \tnode{$\lambda_3$}, \tnode{$2$} ) \right)\\
        &\circ_{z_3} \tnode{$\cdot$} \left( \tnode{\footnotesize pow} ( \tnode{$\lambda_3$}, \tnode{$2$} ), \tnode{\footnotesize pow} ( \tnode{$\lambda_1$}, \tnode{$2$} ) \right)\\
        E_J &:= \tnode{$\cdot$} \left( \tnode{$\cdot$} \left( \tnode{$\lambda_1$}, \tnode{$\lambda_2$} \right), \tnode{$\lambda_3$} \right) \\
        E_\text{MRS} &:= \tnode{$+$} \left( \tnode{$\cdot$} \left( \tnode{$C_1$},
        \tnode{-}\left(
        E_{\Bar{I}_1}
        , \tnode{3} \right) \right), \tnode{$\cdot$} \left( \tnode{$C_2$},
        \tnode{-}\left(
        E_{\Bar{I}_2}
        , \tnode{3} \right) \right) \right)
    \end{aligned}
\end{equation}
with (real) parameters associated with the occurrence of symbols \tnode{$C_1$}, \tnode{$C_2$}

\subsubsection{Generalized Rivlin Model}
\label{sec:general_rivlin}
The previous model can be generalized to
\begin{equation}
    W = \sum_{p,q = 0}^N C_{pq} (\bar{I}_1 - 3)^p \cdot (\bar{I}_2 - 3)^q + \sum_{m=1}^{M} \frac{1}{D_m} (J - 1)^{2m}
\end{equation}
expressed as
\begin{equation}
    \label{eq:mooney_rivlin_prte}
    \begin{aligned}
        E_\text{GRM} &:= \tnode{$+$} \left( \left( \tnode{$+$}(y, x) + y \right)^{\infty x}, \left( \tnode{$+$}(z, x) + z \right)^{\infty x} \right) \\
        &\circ_y \tnode{$\cdot$}\left( \tnode{C}, \left( \tnode{$\cdot$} \left( \tnode{-}(E_{\bar{I}_1}, \tnode{3}), z_q \right) + z_q \right)^{\infty z_p} \circ_{z_q} \left( \tnode{$\cdot$} \left( \tnode{-}(E_{\bar{I}_2}, \tnode{3}), z_q \right) + \tnode{1} \right)^{\infty z_q} \right) \\
        &\circ_z \tnode{$\cdot$}\left( \tnode{D}, \left( \tnode{$\cdot$} \left( \tnode{\footnotesize pow}\left( \tnode{-}(E_{J}, \tnode{1}), \tnode{2} \right), z_m \right) + \tnode{\footnotesize pow}\left( \tnode{-}(E_{J}, \tnode{1}), \tnode{2} \right) \right)^{\infty z_m} \right)
    \end{aligned}
\end{equation}
with (real) parameters for every occurence of \tnode{$C$}, \tnode{$D$} and $E_{\bar{I}_1}, E_{\bar{I}_2}, E_J$ from the previous example.

%% file: sections/appendix/newtonian.tex
In our experiments, we considered the combination of different priors that reflect different knowledge an expert can have about the forces present in a \textit{particle system} derived from Newtonian physics. Each assumption can be independently expressed through an \gls{pRTE} as visualized in \cref{table:newtonian}. The last column shows the combination of knowledge which can be automatically derived using boolean operations. The first row shows the different priors over force functions when used to integrate the dynamics of a test particle system. The second row is a short description of the knowledge that is contained in the prior, followed by a corresponding \gls{pRTE}. Note that we reduced the search space to be able to visualize the resulting \gls{PTA} in the next row. All experiments in the main paper do \textit{not} use this version but the full search space and knowledge as described in \citet{xu2021bayesian}. For the visualization of the state machines we follow the convention from \citet{weidner2015probabilistic} and depict states as nodes, and an arrow between nodes iff the transition probability after reading a symbol $s \in \Sigma^{(1)}$ is greater than zero. For symbols $s \in \Sigma^{(2)}$ of rank two, there are two consecutive states and thus the solid arrow depicts the transition for the left child, while the dashed arrow depicts the transition for the right child. Outgoing arrows depict the final states. Recall that the automaton accepts when there is a run that transitions to only final states in the leaf nodes.

\begin{table*}
    \caption[short]{Newtonian Physics expressed as regular tree priors}
    \label{table:newtonian}
    \vskip 0.15in
    \centering
    \begin{small}
        \begin{center}
            \begin{tabular}{p{0.5cm} m{3.75cm} m{3.75cm} m{3.75cm}}
                \toprule
                & Translational Invariance & Dimensional Analysis & Combined Knowledge \\
                \midrule
                Prior&
                \includegraphics*[width=3.73cm]{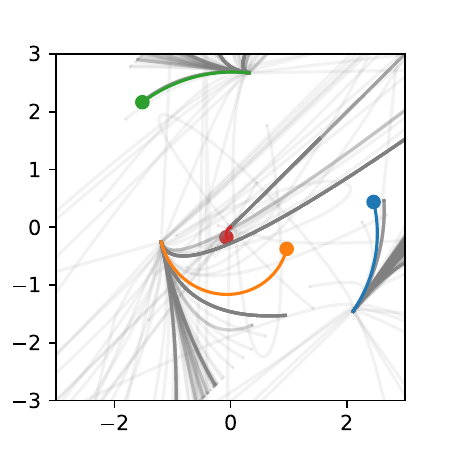} &
                \includegraphics*[width=3.73cm]{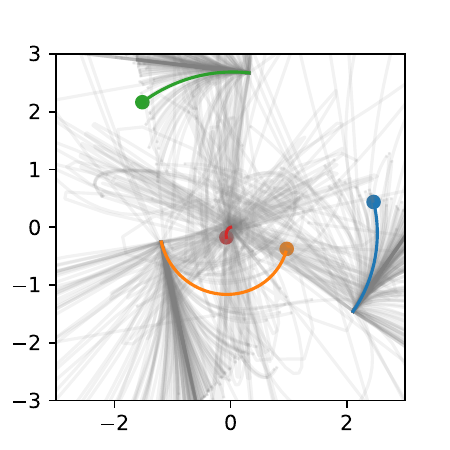} & 
                \includegraphics*[width=3.73cm]{figures/newtonian/prior_sample_star_system_2027.pdf} \\
                Desc. &
                Positions $p_i, p_j$ are only allowed to appear as differences &
                E.g. only quantities of matching units (e.g.\ $\text{kg}$, $\text{kg}^2$) are allowed to be added. (The same holds true for other units and unitless vectors) &
                A combination of arbitrary equations that have both properties. \\
                \\
                pRTE &
                \rotatebox{90}{
                $\begin{aligned}
                    &\ldots \\
                    &\left( \tnode{+}(x,x) + \tnode{$-$}(x,x) + {\scriptstyle \ldots} + y \right)^{\infty x}\\
                    &\circ_y \left( \tnode{$-$}(\tnode{$p_i$}, \tnode{$p_j$}) + \tnode{$-$}(\tnode{$p_j$}, \tnode{$p_i$}) \right) \\
                    &\ldots
                \end{aligned}$
                }&
                \rotatebox{90}{
                $\begin{aligned}
                    \ldots \hspace{10em}\\
                    E_{kg2} = \left( \tnode{$\cdot^2$}(E_{kg}) + \tnode{$\cdot$}(\tnode{$m_i$}, \tnode{$m_j$})\right) \\
                    E_{kg} = \left( 
                    \begin{aligned}
                        \tnode{$m_i$} + \tnode{$m_j$} + \\
                        \tnode{$-$}(\tnode{$m_i$}, \tnode{$m_j$}) + \\
                         \tnode{$+$}(\tnode{$m_i$}, \tnode{$m_j$})
                    \end{aligned} 
                    \right) \hspace{2em} \\
                    \ldots \hspace{10em}
                \end{aligned}$
                }&
                \begin{equation*}E_\textit{inv} \cap E_\text{dim} \end{equation*}\\
                \\
                PTA &
                \scalebox{0.8}{\input{figures/pta/transinv_pta.tex}} &
                \scalebox{0.8}{\input{figures/pta/diman_pta.tex}} &
                \scalebox{0.8}{\input{figures/pta/both_pta.tex}} \\
                \bottomrule
            \end{tabular}
        \end{center}
    \end{small}
\end{table*}

%% file: figures/pta/transinv_pta.tex
\begin{tikzpicture}
    \node[draw, circle] (q0) {$q_0$};
    \node[draw, circle, right= 1.5cm of q0, yshift=-0.5cm] (q1) {$q_1$};
    \node[draw, circle, right= 1.5cm of q0, yshift=0.5cm] (q2) {$q_2$};

    \node[left= 9mm of q0] (out0) {};
    \node[right= 5mm of q1] () (out1) {};
    \node[right= 5mm of q2] () (out2) {};

    \draw[arrows = {Circle[]-Computer Modern Rightarrow[line cap=round]}, black] (q0) to[out=90, in=120, loop, looseness=10] node[midway, above]{\tiny $+, -, \ldots$} (q0);
    \draw[arrows = {-Computer Modern Rightarrow[line cap=round]}, black, dashed] (q0) to[out=90, in=60, loop, looseness=10] (q0);

    \draw[arrows = {Circle[]-Computer Modern Rightarrow[line cap=round]}, black] (q0) to[out=-90, in=-120, loop, looseness=10] node[midway, below]{\tiny $\exp, \sin, \ldots$} (q0);

    \draw[arrows = {Circle[]-Computer Modern Rightarrow[line cap=round]}, black] (q0) to[out=-45, in=-160] node[midway, below]{\tiny $+, -, \ldots$} (q1);
    \draw[arrows = {-Computer Modern Rightarrow[line cap=round]}, black, dashed] (q0) to[out=-45, in=-160] (q2);

    \draw[arrows = {Circle[]-Computer Modern Rightarrow[line cap=round]}, black] (q0) to[out=30, in=-200] node[midway, above]{\tiny $+, -, \ldots$} (q2);
    \draw[arrows = {-Computer Modern Rightarrow[line cap=round]}, black, dashed] (q0) to[out=30, in=-200] (q1);

    \draw[arrows = {-Computer Modern Rightarrow[line cap=round]}, black] (q0) -- node[midway, above] {\tiny $m_i, m_j$} (out0);
    \draw[arrows = {-Computer Modern Rightarrow[line cap=round]}, black] (q1) -- node[midway, above] {\tiny $p_i$} (out1);
    \draw[arrows = {-Computer Modern Rightarrow[line cap=round]}, black] (q2) -- node[midway, above] {\tiny $p_j$} (out2);
\end{tikzpicture}

%% file: figures/pta/diman_pta.tex
\begin{tikzpicture}
    \node[draw, circle] (q0) {$q_0$};
    \node[draw, circle, above= 1cm of q0] (q4) {$q_4$};
    \node[draw, circle, right= 2cm of q4] (q3) {$q_3$};
    \node[draw, circle, right= .5cm of q0, yshift=0.25cm] (q1) {$q_1$};
    \node[draw, circle, right= .5cm of q1, yshift=0.25cm] (q2) {$q_2$};

    \node[below= 3mm of q1] (out1) {};
    \node[below= 3mm of q2] (out2) {};
    \node[right= 9mm of q3] (out3) {};

    \draw[arrows = {Circle[]-Computer Modern Rightarrow[line cap=round]}, black] (q0) to[out=90, in=-60] node[midway, above]{\tiny $+, -$} (q4);
    \draw[arrows = {-Computer Modern Rightarrow[line cap=round]}, black, dashed] (q0) to[out=90, in=-120] (q4);


    \draw[arrows = {Circle[]-Computer Modern Rightarrow[line cap=round]}, black] (q4) to[out=-5, in=120] node[midway, right]{\tiny $\times$} (q1);
    \draw[arrows = {-Computer Modern Rightarrow[line cap=round]}, black, dashed] (q4) to[out=-5, in=140] (q2);

    \draw[arrows = {Circle[]-Computer Modern Rightarrow[line cap=round]}, black] (q4) to[out=20, in=160] node[midway, above]{\tiny $\cdot^2$} (q3);
    
    \draw[arrows = {Circle[]-Computer Modern Rightarrow[line cap=round]}, black] (q3) to[out=-140, in=70] node[midway, right]{\tiny $+,-$} (q1);
    \draw[arrows = {-Computer Modern Rightarrow[line cap=round]}, black, dashed] (q3) to[out=-140, in=45] (q2);

    \draw[arrows = {-Computer Modern Rightarrow[line cap=round]}, black] (q1) -- node[midway, right] {\tiny $m_i$} (out1);
    \draw[arrows = {-Computer Modern Rightarrow[line cap=round]}, black] (q2) -- node[midway, right] {\tiny $m_j$} (out2);
    \draw[arrows = {-Computer Modern Rightarrow[line cap=round]}, black] (q3) -- node[midway, above] {\tiny $m_i, m_j$} (out3);
\end{tikzpicture}

%% file: figures/pta/both_pta.tex
\begin{tikzpicture}
    \node[draw, circle] (q0) {$q_0$};
    \node[draw, circle, right= 0.75cm of q0, yshift=-0.5cm] (q4) {$q_4$};
    \node[draw, circle, right= 2cm of q4] (q3) {$q_3$};
    \node[draw, circle, below= .75cm of q4, xshift=0.75cm] (q1) {$q_1$};
    \node[draw, circle, right= .5cm of q1, yshift=0.25cm] (q2) {$q_2$};

    \node[below= 3mm of q1] (out1) {};
    \node[below= 3mm of q2] (out2) {};
    \node[below= 3mm of q3] (out3) {};

    \draw[arrows = {Circle[]-Computer Modern Rightarrow[line cap=round]}, black] (q0) to[out=0, in=-120] node[midway, left]{\tiny $+, -$} (q4);
    \draw[arrows = {-Computer Modern Rightarrow[line cap=round]}, black, dashed] (q0) to[out=0, in=-200] (q4);


    \draw[arrows = {Circle[]-Computer Modern Rightarrow[line cap=round]}, black] (q4) to[out=-5, in=120] node[midway, right]{\tiny $\times$} (q1);
    \draw[arrows = {-Computer Modern Rightarrow[line cap=round]}, black, dashed] (q4) to[out=-5, in=140] (q2);

    \draw[arrows = {Circle[]-Computer Modern Rightarrow[line cap=round]}, black] (q4) to[out=20, in=160] node[midway, below]{\tiny $\cdot^2$} (q3);
    
    \draw[arrows = {Circle[]-Computer Modern Rightarrow[line cap=round]}, black] (q3) to[out=-140, in=70] node[midway, above right]{\tiny $+,-$} (q1);
    \draw[arrows = {-Computer Modern Rightarrow[line cap=round]}, black, dashed] (q3) to[out=-140, in=45] (q2);

    \draw[arrows = {-Computer Modern Rightarrow[line cap=round]}, black] (q1) -- node[midway, right] {\tiny $m_i$} (out1);
    \draw[arrows = {-Computer Modern Rightarrow[line cap=round]}, black] (q2) -- node[midway, right] {\tiny $m_j$} (out2);
    \draw[arrows = {-Computer Modern Rightarrow[line cap=round]}, black] (q3) -- node[midway, right] {\tiny $m_i, m_j$} (out3);

    \node[draw, circle, above= 1cm of q4] (q5) {$q_5$};
    \node[draw, circle, right= 1.5cm of q5, yshift=-0.5cm] (q6) {$q_6$};
    \node[draw, circle, right= 1.5cm of q5, yshift=0.5cm] (q7) {$q_7$};

    \node[left= 9mm of q5] (out5) {};
    \node[right= 5mm of q6] () (out6) {};
    \node[right= 5mm of q7] () (out7) {};

    \draw[arrows = {Circle[]-Computer Modern Rightarrow[line cap=round]}, black] (q5) to[out=90, in=120, loop, looseness=10] node[midway, above]{\tiny $+, -, \ldots$} (q5);
    \draw[arrows = {-Computer Modern Rightarrow[line cap=round]}, black, dashed] (q5) to[out=90, in=60, loop, looseness=10] (q5);


    \draw[arrows = {Circle[]-Computer Modern Rightarrow[line cap=round]}, black] (q5) to[out=-45, in=-160] node[midway, above]{\tiny $+, -, \ldots$} (q6);
    \draw[arrows = {-Computer Modern Rightarrow[line cap=round]}, black, dashed] (q5) to[out=-45, in=-160] (q7);

    \draw[arrows = {Circle[]-Computer Modern Rightarrow[line cap=round]}, black] (q5) to[out=30, in=-200] node[midway, above]{\tiny $+, -, \ldots$} (q7);
    \draw[arrows = {-Computer Modern Rightarrow[line cap=round]}, black, dashed] (q5) to[out=30, in=-200] (q6);

    \draw[arrows = {Circle[]-Computer Modern Rightarrow[line cap=round]}, black] (q0) to[out=0, in=-100] node[midway, above]{\tiny $+, -$} (q5);
    \draw[arrows = {-Computer Modern Rightarrow[line cap=round]}, black, dashed] (q0) to[out=0, in=-150] (q5);

    \draw[arrows = {Circle[]-Computer Modern Rightarrow[line cap=round]}, black] (q0) to[out=90, in=-180] node[midway, above]{\tiny $\exp, \sin, \ldots$} (q5);

    \draw[arrows = {-Computer Modern Rightarrow[line cap=round]}, black] (q6) -- node[midway, above] {\tiny $p_i$} (out6);
    \draw[arrows = {-Computer Modern Rightarrow[line cap=round]}, black] (q7) -- node[midway, above] {\tiny $p_j$} (out7);
\end{tikzpicture}

%% file: sections/appendix/complexity.tex
We conducted a \textit{complexity analysis} of the expressions found by the different algorithms to better understand the generalization performance of \textit{pRTE}.
As a measure of the complexity of expressions, we used the idea of \textit{description length} from information theory.
The more bits are required to describe a model, the more complex it is considered.
The same approach is followed by \citet{udrescu2020ai2} and results in the combined description length for both parameters and structure of an equation as detailed in \cref{tab:description_length}.
Symbolic expressions are considered parameterized functions for which each node within its tree structure is considered a basis function.
For instance, the expression $y \cdot (2 \cdot x + 7)$ has a description length of $\mathcal{L}_D = 5 \cdot \log_2(4) + \mathcal{L}_D([2, 7])$ resulting from the $5$ occurrences of the $4$ unique
basis functions $\{ \cdot, +, x, y\}$.
\begin{table}
    \centering
    \caption{Defintion of \textit{description length} for symbolic expressions following \citet{udrescu2020ai2}.}
    \label{tab:description_length}
    \vskip 0.15in
    \begin{tabular}{ccc}
        \hline
        Object & Symbol & Description length $\mathcal{L}_D$\\
        \hline
        Natural Number & $n$ & $\log_2(n)$ \\
        Integer & $m$ & $\log_2(1 + |m|)$ \\
        Rational Number & $\frac{m}{n}$ & $\mathcal{L}_D(m) + \mathcal{L}_D(n)$ \\
        Real number & r & $\log_+(\frac{r}{\epsilon}), \;\;\; \log_+(x) := \frac{1}{2} \log_2(1 + x^2)$ \\
        Parameter vector & $\mathbf{p}$ & $\sum_i \mathcal{L}_D(p_i)$ \\
        Parameterized function & $f(x; \mathbf{p})$ & $\mathcal{L}_D(\mathbf{p}) + k \log_2 n; \;\; \text{$n$ basis functions appeared $k$ times}$ \\
        \hline
    \end{tabular}
    
\end{table}

\begin{table}
    \centering
    \caption{Expected \textit{description lengths} for different problems and algorithms.}
    \label{tab:model_complexities}
    \vskip 0.15in
    \input{tables/complexity/all_complexities.tex}
\end{table}

\begin{figure}
    \includegraphics[width=\textwidth]{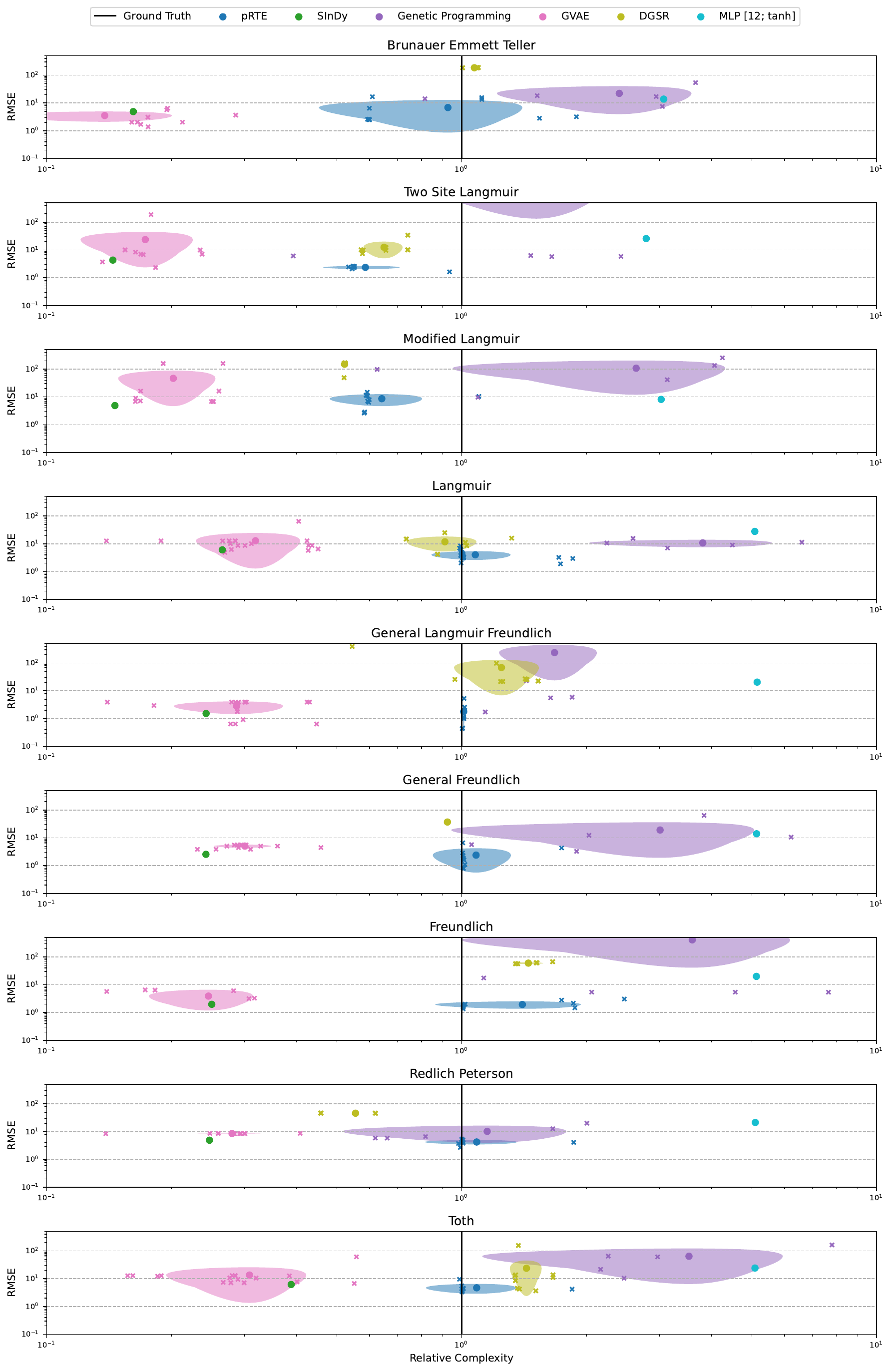}
    \caption{Complexity analysis of the found symbolic expressions for different problems and algorithms: 
    solid dots indicate the expected relative complexity compared to the average prediction performance (RMSE) on the out-of-distribution test set $\mathcal{D}_{test2}$. 
    Shaded areas visualize the corresponding standard deviations. The ground truth model (vertical black line) has a relative complexity of $1.0$. 
    \gls{pRTE} yields solutions that are closest to this ideal complexity and thus generalize better.}
    \label{fig:complexity_analysis}
\end{figure}

The results are summarized in \cref{fig:complexity_analysis}. 
The scatters show the performance of several equations found by the different algorithms in terms of \textit{relative complexity} $\frac{\hat{\mathcal{L}_D}}{\mathcal{L}_D}$ and \textit{prediction error (RMSE)} on an out-of-distribution dataset ($D_\text{test2}$).
Here, $\hat{\mathcal{L}_D}$ is the description length of the symbolic expression found by the algorithm and $\mathcal{L}_D$ is the description length of the ground truth model the data was generated from.
The vertical lines indicate the mean relative complexity of expression found by a particular algorithm while the highlighted areas are a standard deviation to it.

Throughout all different setups \gls{pRTE} is closest to a relative complexity of $1$. 
This means that the expert knowledge in the tree priors helps to better find the right level of complexity of expression for the problem which we hypothesize leads to \gls{pRTE}'s superior prediction performance
in out-of-distribution datasets.
While the implicit priors encoded to \textit{GVAE, SInDy} lead to too simple expressions, the expressions found by genetic programming are often close in complexity to a small neural network (muli-layer perceptron with $12$ hidden units and $\tanh$ activations).

%% file: tables/complexity/all_complexities.tex
\centering
{\def\arraystretch{3}
\begin{tabular}{ccrrrrrrr}
    \toprule
    \vspace{-1em}
    & \multicolumn{7}{c}{Description length $\mathbb{E}[\mathcal{L}_D]$ [bits]} \\
    \textbf{Sorption Isotherm} & True & pRTE & GP & SInDy & NN & GVAE & DGSR \\
    \midrule
    \shortstack[c]{Brunauer Emmett Teller \\ $\frac{k_1 c}{1 + k_2 c} \frac{1}{1 - k_3 c}$} & 756.9 & 701.1 & 1814.5 & 122.4 & 2322.4 & 104.4 & 812.1 \\
    
    \shortstack[c]{Two Site Langmuir \\ $s_T \left( \frac{f_1 k_1 c}{1 + k_1 c} + \frac{f_2 k_2 c}{1 + k_2 c} \right)$} & 829.9 & 486.4 & 1256.4 & 119.8 & 2310.6 & 143.5 & 539.5 \\
    
    \shortstack[c]{Modified Langmuir \\ $s_T\frac{k_1 c}{1 + k_1 c} \frac{1}{1 + k_2 c}$} & 768.2 & 493.2 & 2022.4 & 112.2 & 2323.8 & 155.1 & 401.1 \\
    
    \shortstack[c]{Langmuir \\ $s_T\frac{kc}{1 + kc}$} & 454.6 & 490.3 & 1732.2 & 120.4 & 2311.8 & 144.8 & 414.1 \\
    
    \shortstack[c]{General Langmuir Freundlich \\ $s_T\frac{(kc)^\alpha}{1 + (kc)^\alpha}$} & 450.1 & 455.1 & 753.5 & 109.0 & 2318.4 & 129.1 & 561.2\\
    
    \shortstack[c]{Farley Dzombak Morel \\ \scalebox{0.65}{ $\frac{s_T k_1 c}{1 + k_1 c} + \left(\frac{X - s_T}{1 + k_1 c} + \frac{k_1 X_c}{1 + k_1 c} \right) \times \left( \frac{k_2 c}{1 - k_2 c} \right) - \frac{k_2}{k_3}c$}} & 1763.3 & 517.3 & 1069.4 & 118.7 & 2314.3 & 143.4 & 700.3 \\
    
    \shortstack[c]{General Freundlich \\ $s_T\left(\frac{kc}{1 + kc}\right)^\alpha$} & 450.1 & 487.5 & 1353.2 & 108.9 & 2312.0 & 135.2 & 415.8 \\
    
    \shortstack[c]{Freundlich \\ $K_F c^\alpha$} & 451.1 & 631.6 & 1621.3 & 112.7 & 2314.6 & 110.7 & 653.0 \\
    
    \shortstack[c]{Redlich Peterson \\ $s_T\frac{kc}{1 + (kc)^\alpha}$} & 454.9 & 494.5 & 524.0 & 112.2 & 2320.1 & 127.2 & 252.4 \\
    
    \shortstack[c]{Toth \\ $s_T\frac{kc}{(1 + (kc)^\alpha)^{1/\alpha}}$} & 454.9 & 494.4 & 1605.5 & 176.6 & 2315.4 & 140.1 & 651.7 \\
    
    \bottomrule
\end{tabular}}